\documentclass{bmvc2k}

\title{CLIP Adaptation by Intra-Modal Overlap Reduction} 

\addauthor{Alexey Kravets}{ak3095@bath.ac.uk}{1}
\addauthor{Vinay Namboodiri}{vpn22@bath.ac.uk}{1}
\addinstitution{
 Department of Computer Science \\
 University of Bath\\
 Bath, UK
}
\usepackage{graphicx,wrapfig}
\usepackage{booktabs}

\usepackage{makecell}
\usepackage{caption}
\usepackage{array}
\usepackage{multirow}
\usepackage{subcaption}
\usepackage{amssymb}

\runninghead{Kravets et. al}{CLIP Adaptation by Intra-Modal Overlap Reduction}

\begin{document}

\maketitle

\begin{abstract}
Numerous methods have been proposed to adapt a pre-trained foundational CLIP model for few-shot classification. As CLIP is trained on a large corpus, it generalises well through adaptation to few-shot classification. In this work, we analyse the intra-modal overlap in image space in terms of embedding representation. Our analysis shows that, due to contrastive learning, embeddings from CLIP model exhibit high cosine similarity distribution overlap in the image space between paired and unpaired examples affecting the performance of few-shot training-free classification methods which rely on similarity in the image space for their predictions. To tackle intra-modal overlap we propose to train a lightweight adapter on a generic set of samples from the Google Open Images dataset demonstrating that this improves accuracy for few-shot training-free classification. We validate our contribution through extensive empirical analysis and demonstrate that reducing the intra-modal overlap leads to a) improved performance on a number of standard datasets, b) increased robustness to distribution shift and c) higher feature variance rendering the features more discriminative for downstream tasks.
\end{abstract}

\section{Introduction}

\begin{figure}[!t]
    \centering
    \includegraphics[width=0.7\linewidth]{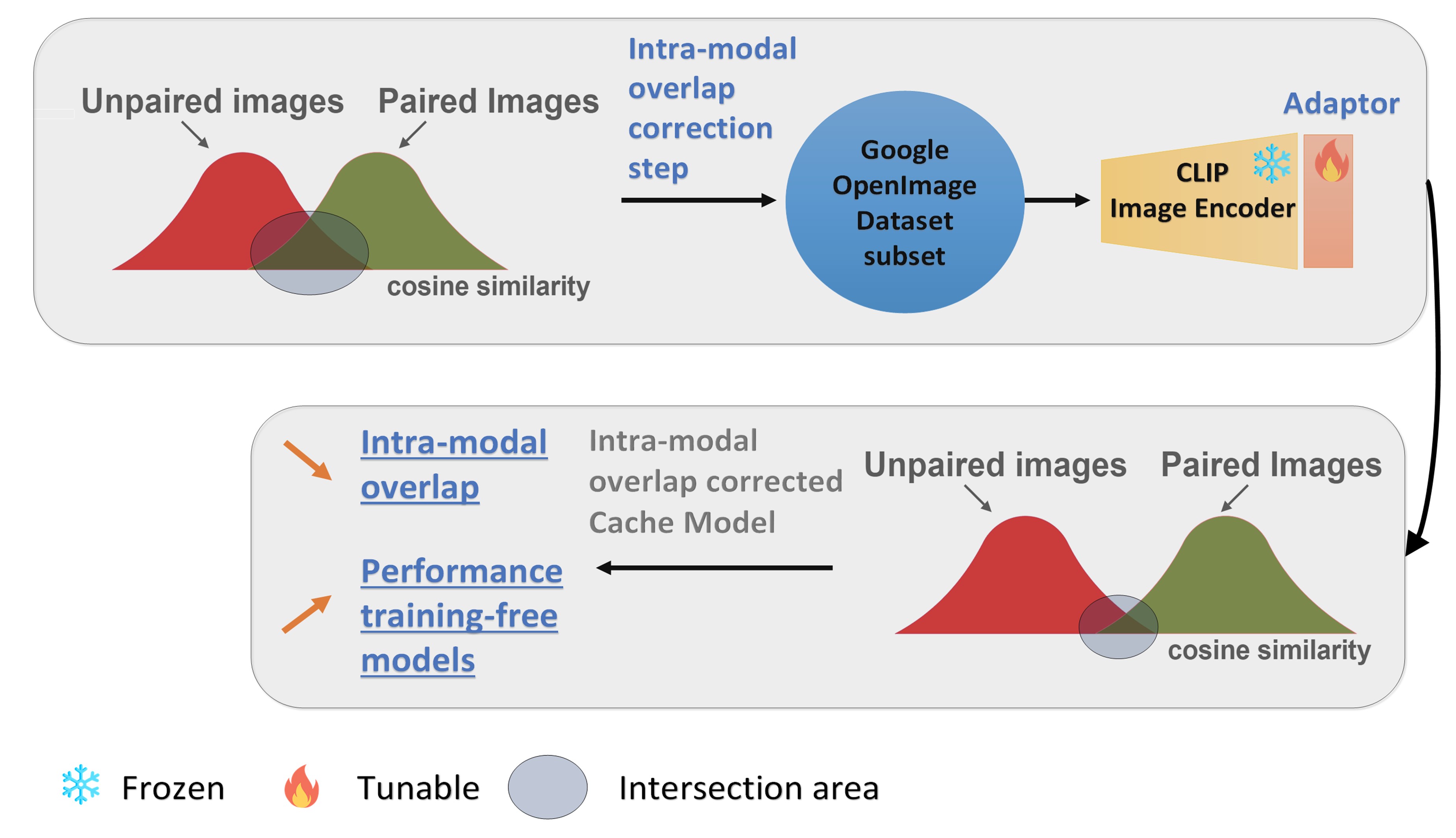}
    \vspace{2mm}
    \caption{Overview of the process. First, we perform a intra-modal overlap correction step of CLIP image encoder through adaptation. Then, this new image encoder is used to create intra-modal overlap corrected cache model that can be used in any training-free method improving its performance. }
    \label{fig:fig_method}
\end{figure}

Vision-language models (VLMs) represent a novel approach in artificial intelligence integrating the understanding of both visual and textual information. An exemplary model within VLMs is CLIP ~\cite{radford_2021_learning}. The fundamental strategy with the advent of large foundation models is to train models with a large number of parameters using vast amounts of data. The training of CLIP follows the same approach. Its task involves learning to match images with their corresponding textual descriptions through contrastive learning. This strategy has resulted in exceptional performance in zero-shot scenarios and requires minimal adaptation across various tasks including semantic segmentation ~\cite{zhou_zegclip}, out-of-distribution detection ~\cite{miyai_2023_locoop}, and classification ~\cite{gao_2021_clipadapter,yu_2023_task}. However, when we try to use this foundational model on a dataset whose distribution is significantly different from the pre-trained data, we observe that the performance is not so good. For instance, zero-shot classification performance of CLIP model on EuroSAT dataset is only 48.38\%. To address this the conventional solution involves collecting a training dataset. However, collecting a large training dataset is often impractical and expensive leading to a collection of only a few examples per class. As few examples are available some approaches ~\cite{gao_2021_clipadapter,yu_2023_task,sun_2023_prompt,zhou_2022_conditional,pantazis_2022_svladapter,zhu_2023_promptaligned,zang_2022_unified,zhou_2022_learning} suggest adapting CLIP by adjusting additional parameters while keeping the original ones frozen. Sometimes, training even small adapters can be infeasible. Thus, alternative approaches ~\cite{udandarao_2023_susx,zhu_2023_not,zhang_2022_tipadapter} propose a training-free adaptation of CLIP. This involves creating a cached model ~\cite{zhang_2022_tipadapter} using CLIP encoded few-shot training images for which labels are available. This cached model can be used to compare a test image to cached images in image space determining the similarity of the test image to few-shot training examples which in combination with zero-shot CLIP logits determines the correct prediction. However, comparing images in the image space with embeddings from CLIP is problematic due to the contrastive training that maximizes the cosine similarity between
paired image and text (inter-modal), but ignores the image-image similarity (intra-modal). This results in a substantial intra-modal overlap (IMO) between unpaired (images of different classes) and paired images (images of the same class) compromising the results of training-free methods that use the cached model. 

We propose a simple approach to address this issue as illustrated in Fig. \ref{fig:fig_method}. The approach is to train a lightweight adapter on a subset of Google Open Images dataset ~\cite{a2019_open} for one epoch. This subset has a different distribution from most of the downstream datasets we test on measured using Proxy-A-Distance \cite{bendavid_2006_analysis} measure of divergence. 
We observe that this simple adaptation step successfully solves the IMO such that the distance between the similarity distributions of paired and unpaired image embeddings successfully increases for many downstream datasets. This approach is thus generalizable and also results in substantially improved performance (for instance performance improvement of around 5\% for one-shot performance on EuroSat dataset taking it to more than 68\% with a single example compared to 48.38\% with zero-shot, cf - detailed table in supplementary material) in many of the downstream datasets. This improvement is complementary to existing approaches and by combining our contribution with \cite{zhang_2022_tipadapter} and \cite{udandarao_2023_susx} results in a consistent overall improvement in accuracy. In this work we mainly focus on fine-grained datasets where the samples are related making classification particularly challenging, but for completeness we perform experiments on some not fine-grained classification datasets whose results will be included in the appendix. 

To summarize, through this paper we make the following \textbf{contributions}: 

\begin{itemize}
    \setlength\itemsep{0.1em}
    \item We propose a novel method based on lightweight adaptation that reduces IMO in CLIP directly in the image space with new features being compatible with any training-free method that utilizes a cached model. These new features improve overall performance in all the training-free methods examined.
    \item We show that there is a positive relation between direct IMO reduction and performance.
    \item We explore the possibility to reduce the IMO by training a lightweight adapter in both supervised and self-supervised manners.
\end{itemize}

\section{Related Work}

\paragraph{Lightweight Adaptation}
Lightweight adaptation is a fine-tuning approach where the majority of parameters in pre-trained models remain fixed and only a small fraction undergoes tuning. While some lightweight adaptation techniques, like prefix-tuning ~\cite{li_2021_prefixtuning}, are specific to Natural Language Processing (NLP), many are versatile and applicable to both NLP and vision models. In ~\cite{houlsby_2019_parameterefficient} authors add sequentially two additional adapter modules inserted in each transformer layer after the projection following the Multi-Head Attention (MHA) layer and the second Multilayer Perceptron (MLP) layer. Each adapter comprises a bottleneck MLP with non-linearity and a residual connection. \cite{pfeiffer_2021_adapterfusion} simplify it further by inserting bottleneck adapter only after the second MLP layer, specifically after the LayerNorm. Low-Rank Adaptation (LoRA) \cite{hu_2021_lora} utilizes low-rank factorization to fine-tune attention weights, significantly reducing the number of parameters during adaptation. AdaptFormer \cite{chen_2022_adaptformer} introduces a bottleneck MLP layer after the MHA of a transformer layer. This newly added MLP layer is parallel with the original MLP and the two are connected via a residual connection with a scale factor. \\
In this study we utilize adapters not for a downstream task adaptation but specifically to address IMO. Furthermore, our focus is on vision adaptation for CLIP vision encoder which is affected by IMO. We are not interested in reducing the intra-modal overlap in text space as text to text matching is not utilized to perform few-shot classification. 

\paragraph{Few-shot Classification Methods}
We can categorize methods utilizing CLIP for few-shot classification into three different groups. Firstly, there are methods like ~\cite{gao_2021_clipadapter,yu_2023_task,sun_2023_prompt,zhou_2022_conditional,pantazis_2022_svladapter,zhu_2023_promptaligned,zang_2022_unified,zhou_2022_learning} that involve training. These methods use few-shot examples to adjust additional parameters while keeping the original CLIP parameters fixed. Secondly, there are zero-shot methods, such as ~\cite{guo_2022_calip,radford_2021_learning}, which do not introduce any extra parameters to CLIP and do not necessitate training. Lastly, there are training-free methods or hybrid methods that are training-free but also might have a training counterpart. In this work we specifically focus on training-free methods ~\cite{zhang_2022_tipadapter,udandarao_2023_susx,zhu_2023_not}, excluding their training counterparts. As all of them utilize the cached model component for prediction which is affected by the IMO ~\cite{udandarao_2023_susx}, we show that replacing it with our IMO corrected cache model component improves performance in all the training-free methods.

\paragraph{Self-supervised Learning in Images}

Self-supervised learning (SSL) involves learning representations from unlabeled data without explicit annotations which is especially valuable when obtaining data labels is costly. While supervised models generally perform better, self-supervised trained models, particularly those based on the contrastive learning paradigm have shown superiority in tasks like segmentation and detection and have been closing the gap in other tasks ~\cite{caron_2021_emerging,he_2020_momentum}. Notable methods include SimCLR ~\cite{chen_2020_a} which relies on contrastive learning and requires a large batch size to incorporate a sufficient number of negative examples, MoCo ~\cite{he_2020_momentum} which utilizes a queue mechanism to store negative samples, and BYOL ~\cite{grill_2020_bootstrap} which introduces a novel paradigm eliminating the need for negative samples. DINO ~\cite{caron_2021_emerging}, like BYOL, relies on positive samples but utilizes cross-entropy loss rather than L2 loss. \\
While SSL methods for training entire networks have been extensively studied there is no exploration training adapters using these methods. We utilize the state-of-the-art DINO method for this purpose and investigate the possibility of training adapters in a self-supervised manner to reduce IMO in CLIP.

\section{Background on Training-free Adaptation}
In this section we provide an overview of training-free adaptation methods for CLIP.
\subsection{Tip-Adapter: the Main Building Component in Training-free Methods}
\label{section:building_comp}

\paragraph{Zero-Shot CLIP}
Given \textit{N} classes, CLIP encodes them inside a contextual prompt such as \textit{A photo of a \{class\}} with the text encoder obtaining $W \in \mathbb{R}^{N\times d}$ classifier weight matrix where \textit{d} is the embedding dimension. Then, given a test image ${I_i}$, it is encoded with CLIP image encoder \textit{VE}:
\begin{equation}
    T_i = VE(I_i), \ T_i \in \mathbb{R}^{d}
\end{equation}
After that, we calculate the dot product between $W$ and $T_i$ to obtain the zero-shot classification logits:
\begin{equation}
    \text{CLIPlogits} = T_i W^T, \ \text{CLIPlogits} \in  \mathbb{R}^{N}
\label{eq:clip_logits}
\end{equation}
\paragraph{Tip-Adapter}
Given \textit{N} classes \textit{K} shots training dataset with images $I_k, k \in \{1, NK\}$, we encode them with CLIP image encoder. Such encoded images act as keys and their corresponding one-hot encoded labels $L_k,  k \in \{1, NK\}$ as values to form the key-value cached model:
\begin{equation}
\begin{split}
& T_k = VE(I_k), k \in [1, NK], T_k \in \mathbb{R}^{d}\\
& F_{train} = \text{Concat}([T_1, T_2, .. , T_{NK}]), F_{train} \in \mathbb{R}^{NK\times d}
\end{split}
\label{eq:ta_features}
\end{equation}
\begin{equation}
\begin{split}
& L_k = \text{OneHot}(L_k), k \in [1, NK], L_k \in \mathbb{R}^{N}\\
& L_{train} = \text{Concat}([L, L_2, .. , L_{NK}]), L_{train} \in \mathbb{R}^{NK\times N}
\end{split}
\label{eq:ta_labels}
\end{equation}
The cached model contains the new knowledge extracted from the few-shot training examples and its purpose is to enhance the prior knowledge of the CLIP model. During the testing phase, when presented with a test image denoted as ${I_i}$, which serves as a query, this image is encoded using the CLIP image encoder \textit{VE}  resulting in a vector representation $T_i \in  \mathbb{R}^{d}$. Subsequently, an affinity matrix is computed. This matrix represents the similarity between the test image and all the \textit{NK} training images: 
\begin{equation}
A = exp(-\beta(1-T_i F_{train}^T)), A \in \mathbb{R}^{NK}
\label{eq:ta_affinity}
\end{equation}
The exponential function makes affinity matrix values non-negative and $\beta$ is a hyper-parameter that modules its sharpness.  \\
After obtaining the affinity matrix and zero-shot CLIP logits we can compute the Tip-Adapter logits by combining the new knowledge of the cached model represented by the product between the affinity matrix and labels matrix $L_{train}$ and the prior knowledge of CLIP:
\begin{equation}
\text{TAlogits} = \alpha A L_{train} + T_i W^T,  \ \text{TAlogits} \in  \mathbb{R}^{N}
\end{equation}
With $\alpha$ being a hyper-parameter that weights the importance of the new and prior knowledge.

\subsection{Tip-X: Inter-modal Bridge for Intra-modal Overlap Correction}

Authors in ~\cite{udandarao_2023_susx} propose to use inter-modal distances as a bridge to handle intra-modal overlap (IMO) between paired and unpaired samples in the image space. They construct an affinity matrix similarly to Tip-Adapter but in the image-text space where the similarity measure between two images is given by Kullback-Leibler (KL) divergence instead of the cosine similarity like in Tip-Adapter. \\
Given test image embedding $T_i \in  \mathbb{R}^{d}$, classifier weight matrix $W \in \mathbb{R}^{N\times d}$,  CLIP encoded few-shot training images $F_{train} \in \mathbb{R}^{NK\times d}$ and their one-hot encoded training labels $L_{train} \in \mathbb{R}^{NK\times N}$ we compute classes probability distribution for train images and the test image:
\begin{equation}
\begin{split}
    & S = \text{SoftMax}(F_{train} W^T), S \in \mathbb{R}^{NK\times N} \\
    & s_i = \text{SoftMax}(T_i W^T), s_i \in \mathbb{R}^{N}
\end{split}
\end{equation}
The affinity matrix $M$ is then constructed by calculating the KL divergence between the test image $s_i$ and the training images $S$. It tells us how closely the distribution of a given test image aligns with the distribution of the training images in the image-text space:
\begin{equation}
    M_{i,j} = KL(s_i || S_j), j \in [1, NK]
\end{equation}
Next, we take the negative of the affinity matrix $M$ because KL divergence is close to 0 for similar images and increases for dissimilar images. It is also rescaled to ensure that it falls within the same range as the Tip-Adapter's affinity matrix. Finally, Tip-X logits are computed by taking the product of the rescaled affinity matrix and the labels matrix $L_{train}$ weighted by a scaler $\gamma$ which is combined with Tip-Adapter logits weighted by a scaler $\alpha$ and CLIP logits to arrive to the final \textit{TXlogits}:
\begin{equation}
\text{TXlogits} =  T_i W^T + \alpha A L_{train} + \gamma \phi(-M) L_{train}, \text{TXlogits} \in  \mathbb{R}^{N}
\end{equation}
While the authors of Tip-X have achieved superior results compared to the original Tip-Adapter, they still incorporate Tip-Adapter logits into the final prediction, which are influenced by the IMO. We later show that replacing this component with IMO-corrected features further improves the results of Tip-X. 

\subsection{Adaptive-Prior Refinement}
A recent work ~\cite{zhu_2023_not} proposes an alternative training-free method to select more discriminative features by eliminating certain feature channels based on a prior refinement module. This method, however, does not reduce the IMO. Hence, we discuss it and provide comparisons in the supplementary material.

\section{Approach}

\subsection{Analysis of Intra-modal Overlap - Intra vs Inter}

\begin{figure}
\centering
\begin{minipage}[b]{0.45\textwidth}
  \centering
  \includegraphics[width=\linewidth]{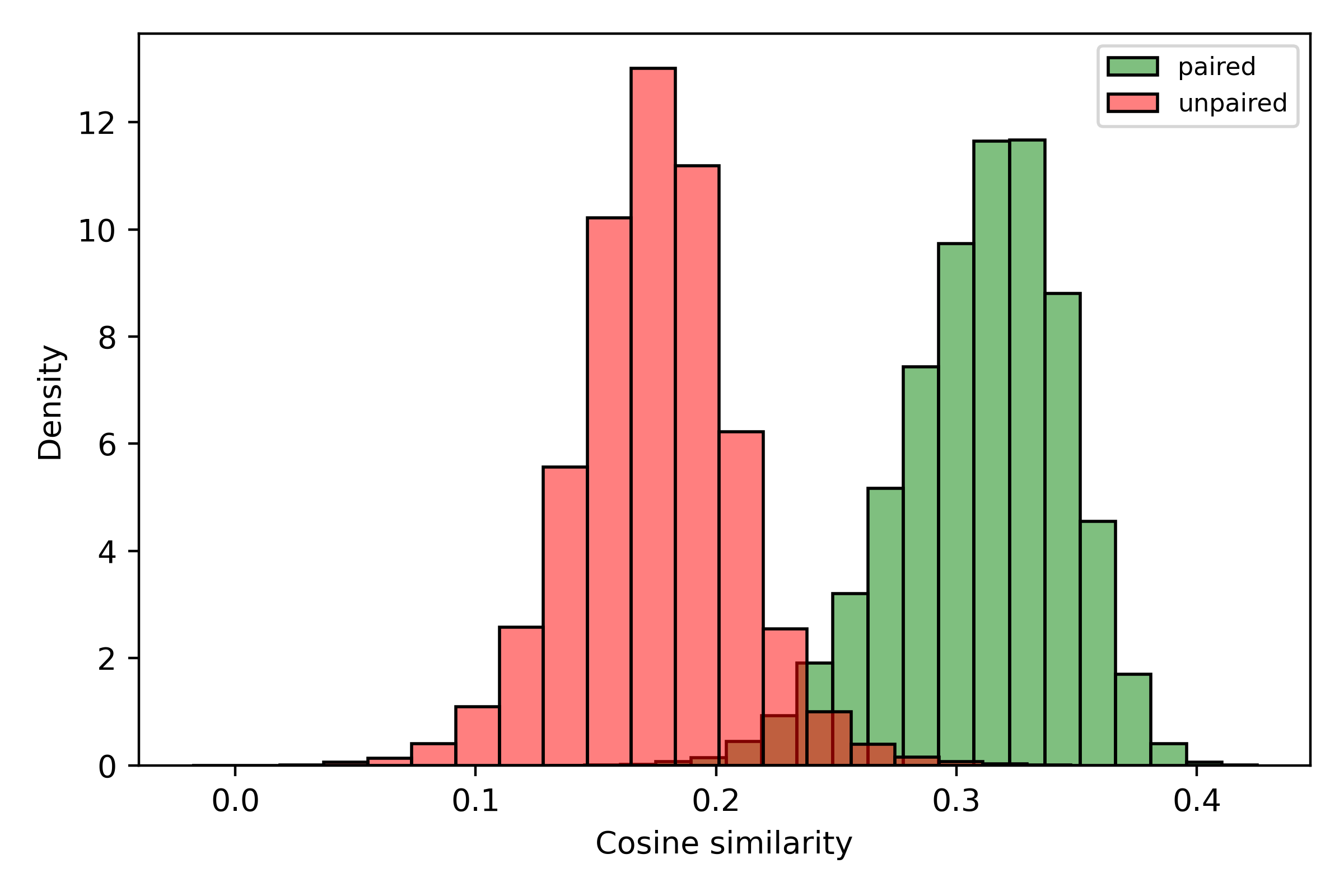}
  \vspace{1mm}
  \caption*{(a) Inter-modal similarity}
\end{minipage}
\begin{minipage}[b]{0.45\textwidth}
  \centering
  \includegraphics[width=\linewidth]{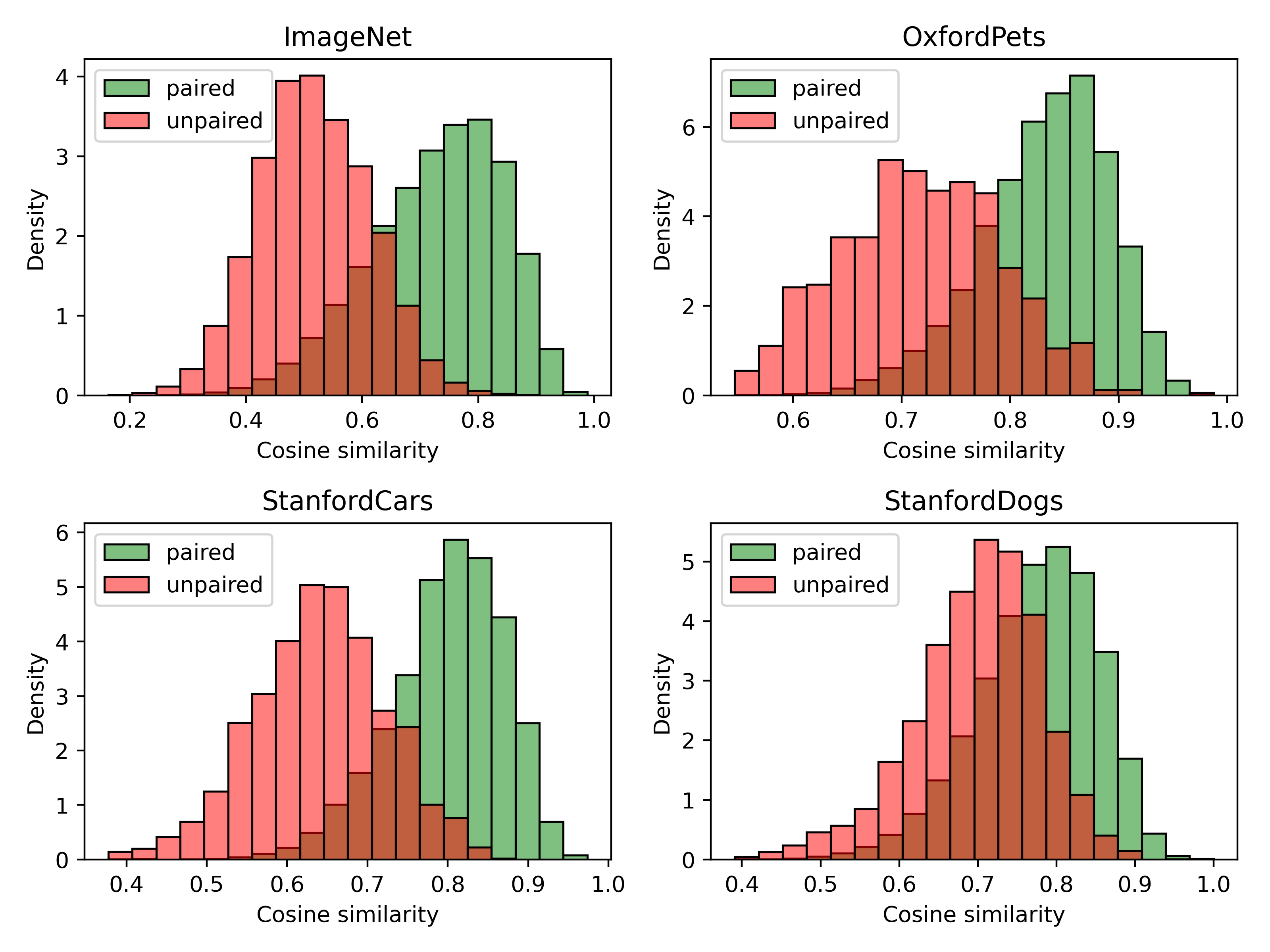}
  \vspace{1mm}
  \caption*{(b) Intra-modal similarity}
\end{minipage}
\vspace*{3mm}
\caption{Fig. (a) shows the inter-modal cosine similarities on the ImageNet validation set. Fig. (b) demonstrates the intra-modal cosine similarities for different datasets on the validation set.}
\label{fig:fig1}
\end{figure}

We analyse the IMO due to contrastive learning that maximizes the cosine similarity between paired image and text (inter-modal) but ignores the image-image similarity (intra-modal) as illustrated in Fig. \ref{fig:fig1}.
We argue that this hampers the performance of few-shot classification. We next proceed to solve this problem.

\subsection{Intra-Modal Overlap Correction via Adaptation}

We provide two methods to correct IMO via adaptation.
\paragraph{Supervised Adapter Fine-tuning}
To correct IMO in CLIP vision encoder we incorporate bottleneck adapters ~\cite{chen_2022_adaptformer} into CLIP visual encoder layers which are fine-tuned in a supervised manner on a small sample of images from Google Open Images dataset (ablations on other standard datasets and number of samples in the Appendix \ref{sec:apppendix_ablations}). Adapters are lightweight components that add 0.80\% (approx. 1M) new parameters to the model with the bottleneck of size 64. All the original weights of CLIP remain frozen.
Following the fine-tuning of CLIP Vision Encoder (\textit{VEimo}) through adapters, we utilize it to create an improved cached model like Tip-Adapter but with IMO-corrected encoded training images $G_{train} \in \mathbb{R}^{NK\times d}$. Then, given a test image encoded with \textit{VEimo}, $U_i \in  \mathbb{R}^{d}$, the affinity matrix \textit{Y} and logits of Tip-Adapter++ (TA++) are calculated as follows:
\begin{equation}
Y = exp(-\beta(1-U_i G_{train}^T)), Y \in \mathbb{R}^{NK}
\label{eq:ta_affinity_modgap}
\end{equation}
\begin{equation}
\text{TA++logits} =  T_i W^T + \alpha YL_{train}, \text{TA++logits}  \in  \mathbb{R}^{N}
\end{equation}
Similarly, we improve standard Tip-X by replacing the Tip-X affinity matrix $A$ with IMO corrected $Y$, obtaining this way Tip-X++ (TX++) logits:
\begin{equation}
\text{TX++logits} =  T_i W^T + \alpha YL_{train} + \gamma \phi(-M) L_{train}, \text{TX++logits} \in  \mathbb{R}^{N}
\end{equation}

Note that when computing CLIP logits in the image-text space we use CLIP without adapters, which are only integrated into CLIP visual encoder when we need to compute similarity in the image space, thus the zero-shot learning capability of the original CLIP model is not affected.

\paragraph{Self-supervised Adapter Fine-tuning via DINO}
We also explore the possibility of training adapters in an unsupervised manner to investigate whether we can reduce the IMO through self-supervised training. While self-supervised methods for training entire neural networks have been extensively studied, there is less exploration into training adapters using these methods. We utilize the state-of-the-art DINO ~\cite{caron_2021_emerging} method for this purpose, although we also experimented with SimCLR ~\cite{chen_2020_a} and BYOL ~\cite{grill_2020_bootstrap} both of which yielded inferior results. We observe that while the self-supervised training method proves effective, it falls short of the supervised alternative. We therefore defer the discussion about the performance and analysis of the same to the supplementary material. 

\section{Experiments - Supervised Training}

\label{section:experiments}

\begin{figure}
\centering
\begin{minipage}[t]{0.45\textwidth}
  \centering
  \includegraphics[width=\linewidth]{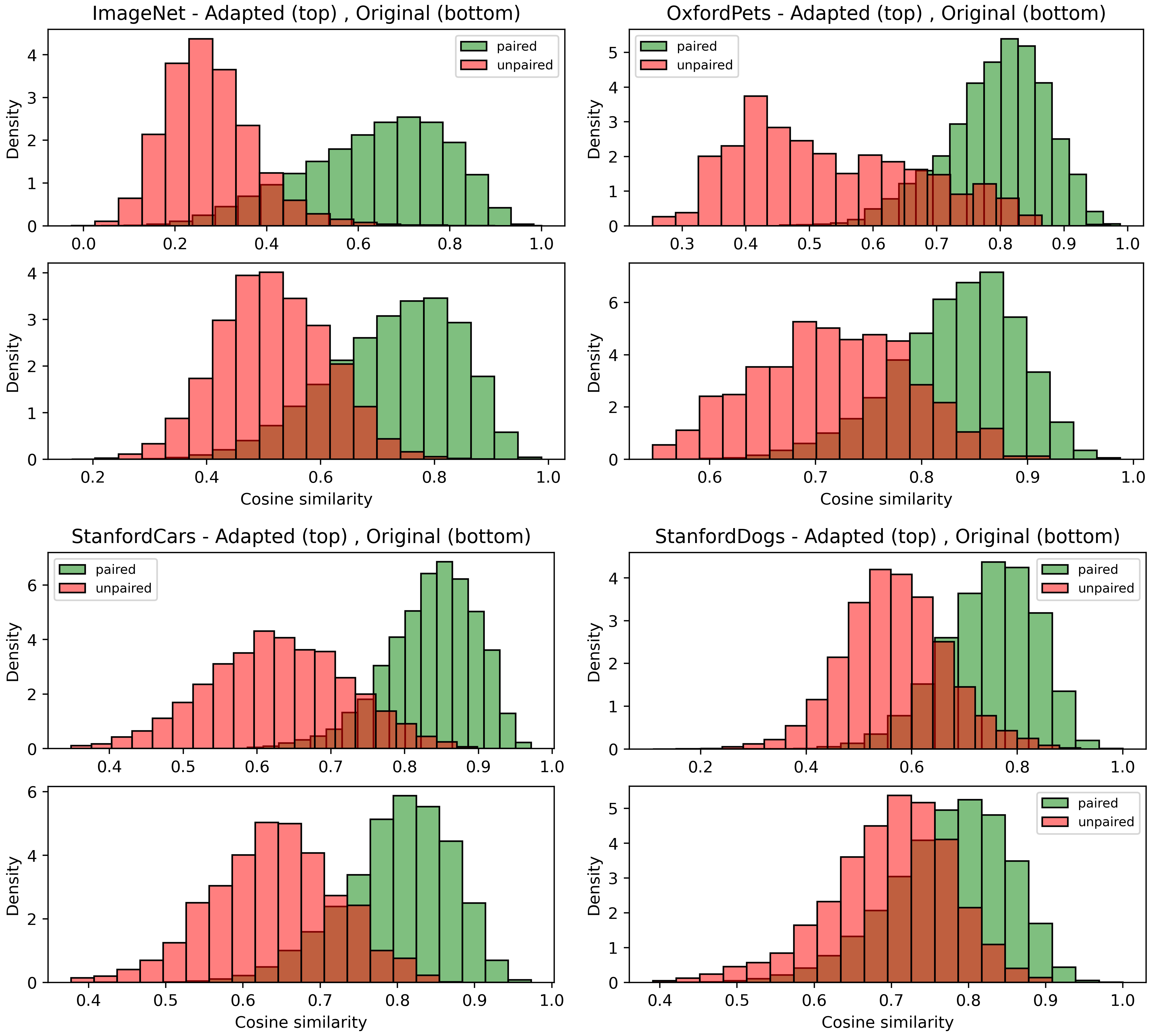}
\end{minipage}%
\begin{minipage}[!t]{0.45\textwidth}
  \centering
  \vspace*{-50mm} % Adjust the vertical space as needed
  \footnotesize
  \begin{tabular}{@{}l ccc@{}}
    \toprule 
    \textbf{Dataset} & \textbf{Adapted} & \textbf{Original} \\
    \midrule 
    ImageNet & 0.1839 & 0.3277 \\
    OxfordPets & 0.3577 & 0.3856 \\
    StanfordCars & 0.2147 & 0.3231 \\
    StanfordDogs & 0.3375 & 0.6208 \\
    \bottomrule
  \end{tabular}
\end{minipage}
\vspace*{3mm}
\caption{Intra-modal overlap measured as intersection area between cosine similarity distribution of paired and unpaired images using adapted and original CLIP image encoder (the lower the better)}
\label{fig:fig2}
\end{figure}

\paragraph{Datasets}
We conduct extensive experiments on 11 fine-grained classification datasets: Caltech101 ~\cite{feifei_2007_learning}, EuroSAT 
~\cite{helber_2019_eurosat}, StanfordCars ~\cite{krause_2013_3d}, OxfordPets ~\cite{parkhi_2012_cats}, DescribableTextures ~\cite{cimpoi_2013_describing}, OxfordFlowers 
~\cite{nilsback_2008_automated}, Food101 ~\cite{bossard_2014_food101}, FGVCAircraft ~\cite{maji_2013_finegrained}, StanfordDogs ~\cite{khosla_novel}, PLANTDOC ~\cite{singh_2020_plantdoc} and CUB ~\cite{he_2020_finegrained}. To ensure completeness, we include results for not fine-grained datasets in some tables. Comprehensive results for not fine-grained datasets will be provided in the supplementary material.

\paragraph{Performance Comparison}

% \begin{wraptable}{r}{9cm}
\begin{table}
    \centering\footnotesize
    \centering
    \fontsize{7}{7}\selectfont
    \setlength{\tabcolsep}{2pt} % Reduce cell padding
    \resizebox{1.\textwidth}{!}{
    \begin{tabular}{@{}lc|cc|cc|cccc@{}} % Adjusted number of columns
    \toprule 
    \textbf{Dataset} & \textbf{Zero-Shot} & \makecell[c]{\textbf{Tip-Adapter}\\ (TA)} & \makecell[c]{\textbf{Tip-Adapter++}\\ (TA++)} & \makecell[c]{\textbf{Tip-X}\\ (TX)} & \makecell[c]{\textbf{Tip-X++}\\ (TX++)} & \makecell[c]{$\Delta$ (TA++, TA)} & \makecell[c]{$\Delta$ (TX++,TX)} & \makecell[c]{$\Delta$ (TA++, TX)}\\
    \midrule 
    EuroSAT & 48.383 & 71.754 & \textbf{74.86} & 71.985 & \textbf{75.364} & 3.106 & 3.379 & 2.875 \\
    StanfordCars & 65.514 & 70.981 & \textbf{73.546} & 73.276 & \textbf{74.744} & 2.565 & 1.467 & 0.27 \\
    PLANTDOC & 34.994 & 47.775 & \textbf{50.25} & 48.206 & \textbf{50.893} & 2.475 & 2.687 & 2.044 \\
    DescribableTextures & 43.972 & 58.676 & \textbf{60.922} & 60.012 & \textbf{61.151} & 2.246 & 1.139 & 0.91 \\
    StanfordDogs & 59.117 & 61.392 & \textbf{63.385} & 64.988 & \textbf{65.438} & 1.993 & 0.45 & -1.603 \\
    SUN397 & 62.579 & 68.746 & \textbf{70.047} & 69.938 & \textbf{70.733} & 1.301 & 0.795 & 0.109 \\
    FGVCAircraft & 24.752 & 33.167 & \textbf{34.401} & 34.945 & \textbf{35.692} & 1.234 & 0.746 & -0.544 \\
    OxfordPets & 89.071 & 90.382 & \textbf{91.567} & 91.569 & \textbf{92.076} & 1.185 & 0.507 & -0.002 \\
    CUB & 55.009 & 65.138 & \textbf{66.042} & 67.088 & \textbf{68.135} & 0.904 & 1.047 & -1.046 \\
    ImageNet & 68.802 & 69.91 & \textbf{70.431} & 70.039 & \textbf{70.468} & 0.521 & 0.429 & 0.392 \\
    Caltech101 & 93.306 & 94.315 & \textbf{94.778} & 94.299 & \textbf{94.799} & 0.462 & 0.5 & 0.479 \\
    Food101 & 85.888 & \textbf{86.195} & 86.165 & 86.253 & \textbf{86.28} & -0.03 & 0.027 & -0.088 \\
    UCF101 & 67.46 & \textbf{75.041} & 74.757 & 76.038 & \textbf{76.098} & -0.284 & 0.06 & -1.281 \\
    OxfordFlowers & 70.767 & \textbf{89.622} & 88.575 & \textbf{90.305} & 89.687 & -1.048 & -0.617 & -1.73 \\
    \midrule
    Average fine-grained & 60.979 & 69.945 & \textbf{71.317} & 71.175 & \textbf{72.205} & 1.372 & 1.03 & 0.142 \\
    Average all & 62.115 & 70.221 & \textbf{71.409} & 71.353 & \textbf{72.254} & 1.188 & 0.901 & 0.056 \\
    \bottomrule
    \end{tabular}}
    \vspace*{2mm}
    \caption{Average performance across all shots on all datasets.}
    \label{table:table1}
\end{table}

Fig. \ref{fig:fig2} illustrates the difference in IMO between the original CLIP visual encoder and the adapted one on the validation set of four different datasets - ImageNet, OxfordPets, StanfordCars and StanfordDogs (the results for all the datasets are in the Appendix \ref{appendix:imo_allds}). The inclusion of the adapter contributes to reducing intra-modal overlap 
between paired and unpaired images. Tab. in Fig. \ref{fig:fig2} quantifies the intersection area between paired and unpaired images (the lower the better). The reduction of IMO is expected to correspond to an improvement in performance. In Tab. \ref{table:table1} we compare the performance of Tip-Adapter and Tip-Adapter++, observing that our method outperforms Tip-Adapter on 11 out of 14 datasets with 1 dataset (Food101) achieving similar results. Additionally, in the same we compare Tip-X and Tip-X++ achieving similar results with Tip-X++ outperforming Tip-X on 13 out of 14 datasets. It is also worth noting that Tip-Adapter++ is competitive or outperforms Tip-X, even with a smaller margin than Tip-X++, on 7 datasets. Overall, Tip-X++ achieves the best performance. These results indicate that our intra-modal overlap corrected encoder is able to extract better features for training-free models. Granular results by number of shots are shown in the Appendix in Fig. \ref{fig:fig4} and Tab. \ref{table:table2} 
where it can be seen that the improvement is usually consistent across different numbers of examples chosen for few-shot classification.

\paragraph{Relation Between Intra-modal Overlap and Performance}

\begin{wrapfigure}{r}{5.5cm}
    \centering\footnotesize
    \includegraphics[width=5.5cm]{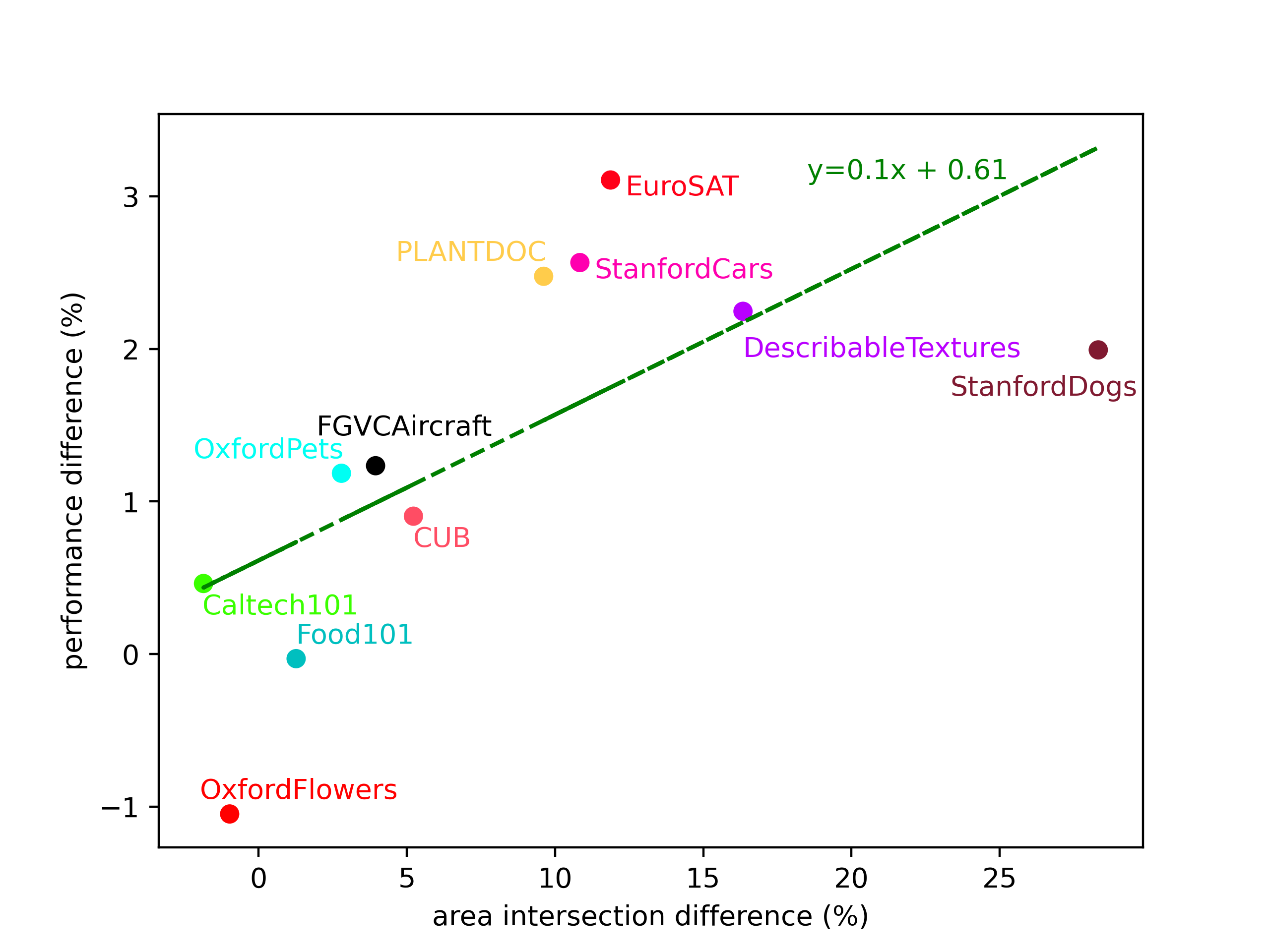}
    \vspace*{2mm}
    \caption{Relation between IMO reduction vs average performance difference between TA++ and TA on fine-grained datasets. }
    \label{fig:fig3}
\end{wrapfigure}

We plot the relation between the difference in intersection area and the average performance difference between Tip-Adapter and Tip-Adapter++. This is to confirm our hypothesis: \textit{the higher the difference in the intersection areas between the original and adapted visual encoders, the higher the performance difference between Tip-Adapter++ and Tip-Adapter} as the IMO reduction was higher. 
This is illustrated in Fig. \ref{fig:fig3} where we observe a positive relation between the two, thus reducing by 1\% the IMO (increasing area intersection difference) leads to approx. 0.10\% improvement of Tip-Adapter++ over Tip-Adapter performance. Furthermore, the two measures exhibit a strong correlation with a correlation coefficient of 0.67. There are, however, few outliers - Food101 has a relatively high difference in intersection areas but the performance of Tip-Adapter++ has not improved over Tip-Adapter. Also, StanfordDogs has a relatively high difference in intersection areas and we expected the performance difference to be higher. 

\paragraph{Robustness to Distribution Shift}

\begin{table}[!t]
\centering
\begin{minipage}{.45\textwidth}
    \centering\footnotesize
    \fontsize{6}{8.9}\selectfont
  \setlength{\tabcolsep}{2pt} % Reduce cell padding
  \begin{tabular}{lccc}
    \toprule 
    \multirow{2}{*}{ \textbf{Models} } & \textbf{Source} & \multicolumn{2}{c}{ \textbf{Target} } \\
    \cline { 2 - 2 } \cline { 3 - 4 }
    & ImageNet & ImageNet-V2 & ImageNet-Sketch \\
    \toprule 
    Zero-Shot CLIP  & 68.804 & 60.83 & 46.14  \\
    \midrule
    Tip-Adapter & 70.753 & 63.02 & 47.24 \\
    Tip-Adapter++ & \textbf{71.505} & \textbf{63.96} & \textbf{48.38} \\
    \midrule
    Tip-X & 70.973 & 63.19 & 47.79 \\
    Tip-X++ & \textbf{71.587} & \textbf{63.98} & \textbf{48.82} \\
    \bottomrule
    \end{tabular}
\vspace{4mm}
\caption{Robustness to distribution shift}
\label{table:table4}
\end{minipage}%
\hspace{5mm} % Add horizontal space between minipages
\begin{minipage}{.5\textwidth}
    \centering\footnotesize
    \fontsize{5}{5}\selectfont
    \setlength{\tabcolsep}{1.5pt} % Reduce cell padding
    \begin{tabular}{@{}lcc@{}} % Adjusted number of columns
    \toprule 
    \textbf{Dataset} & \makecell[c]{\textbf{$\Delta$ (Adapted, Original)}} & \textbf{Proxy-A-Distance} \\
    \midrule 
    CUB & 0.904 & 1.094 \\
    Caltech101 & 0.462 & 0.926 \\
    DescribableTextures & 2.246 & 0.888 \\
    EuroSAT & 3.106 & 1.992 \\
    FGVCAircraft & 1.234 & 1.658 \\
    Food101 & -0.03 & 1.524 \\
    ImageNet & 0.521 & 0.632 \\
    OxfordFlowers & -1.048 & 1.67 \\
    OxfordPets & 1.185 & 1.033 \\
    PLANTDOC & 2.475 & 1.612 \\
    SUN397 & 1.301 & 0.906 \\
    StanfordCars & 2.565 & 1.543 \\
    StanfordDogs & 1.993 & 1.034 \\
    UCF101 & -0.284 & 1.425 \\
    \bottomrule
    \end{tabular}
    \vspace{2mm}
    \caption{Proxy-A-Distance for all datasets.}
    \label{table:table3}
\end{minipage}
\end{table}

We assess the model's robustness to distribution shift. It consists of creating a cached model using one dataset and evaluating it on another. We use ImageNet ~\cite{deng_2009_imagenet} as the source dataset, employing a 16-shot training set, and test on two target datasets: ImageNet-V2 ~\cite{recht_do} and ImageNet-Sketch ~\cite{wang_2019_learning}. These datasets contain similar categories to ImageNet but exhibit semantic gaps. Our findings, shown in Tab. \ref{table:table4} reveal that addressing IMO not only contributes to improved performance when cached model is evaluated on the same dataset but also showcases increased resilience to distribution shift.

\begin{figure}[!t]
\centering
\begin{minipage}{.45\textwidth}
    \centering\footnotesize
    \includegraphics[width=0.7\linewidth]{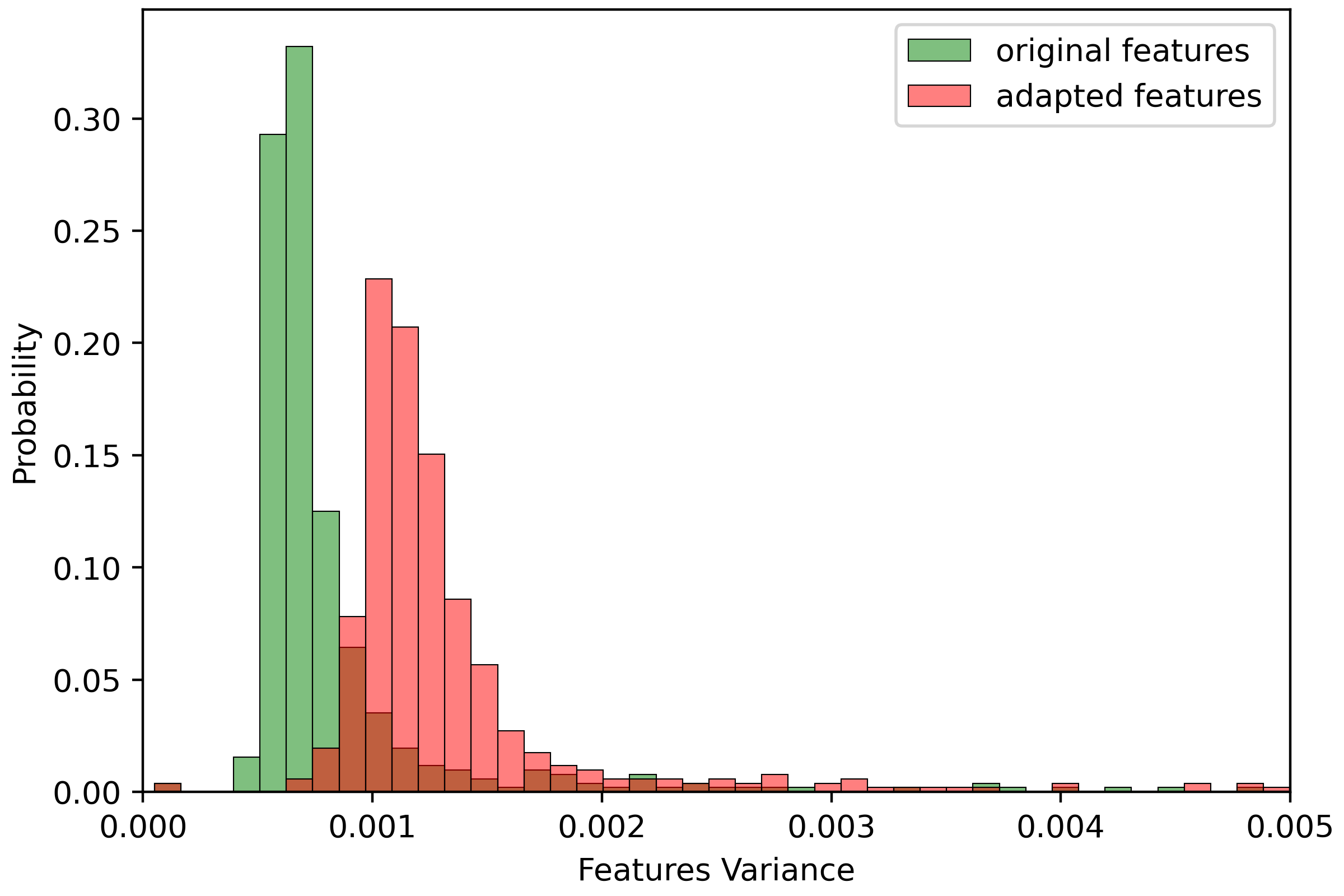}
    \vspace{2mm} % Add vertical space
    \caption{Variance of features on ImageNet validation set of the original and adapted visual encoders.}
    % \vspace{4mm}
    % \caption*{Figure (a)}
    \label{fig:fig5}
\end{minipage}
\hspace{1mm} % Add horizontal space between minipages
\begin{minipage}{.45\textwidth}
    \vspace{6mm}
    \centering
    \includegraphics[width=0.45\linewidth]{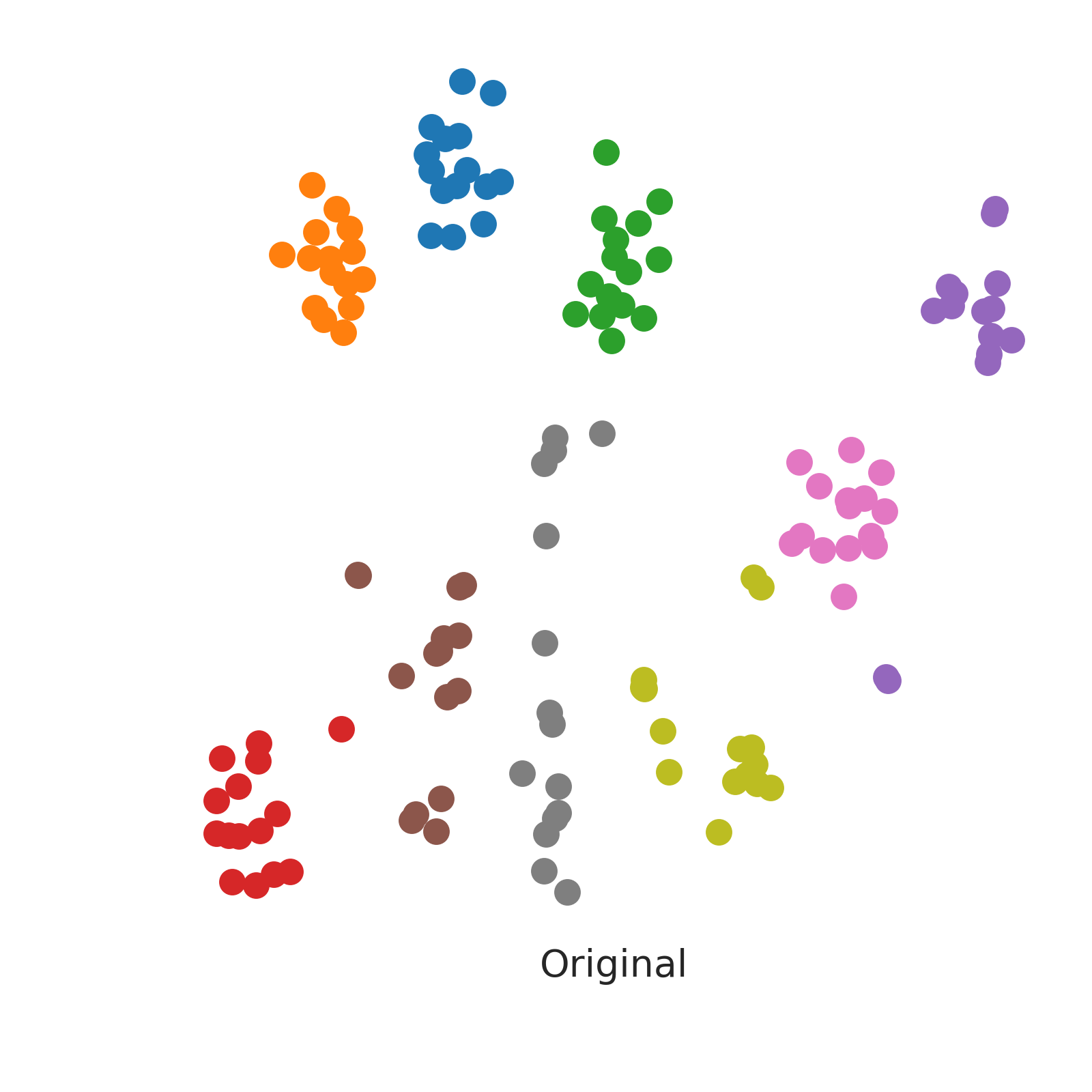} %
    \includegraphics[width=0.45\linewidth]{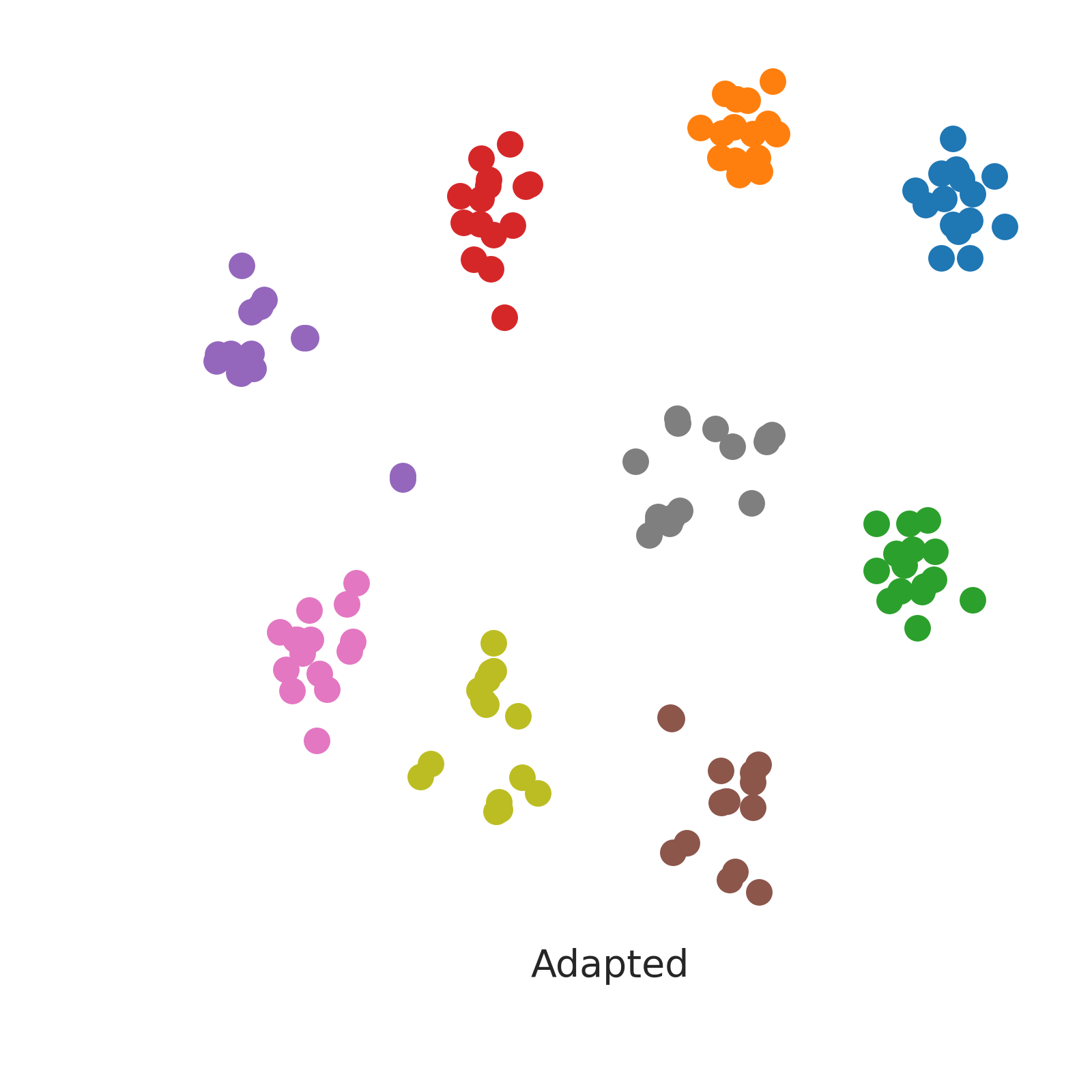} %
    \vspace{1mm}
    % \vspace{4mm}
    % \caption*{Figure (b)}
    \caption{T-SNE visualization of randomly chosen classes from ImageNet validation dataset using original (on the left) and adapted (on the right) visual features.}%
    \label{fig:fig6}%
\end{minipage}
% \vspace{6mm}
\end{figure}

\paragraph{Increase in Features Variance}
We observe that the visual features obtained from CLIP exhibit low variance. Evaluating on ImageNet validation set, as illustrated in Fig. \ref{fig:fig5}, it is apparent that over 50\% of the features exhibit a low variance close to 0. This trend is consistent across all datasets. Low variance across multiple dimensions suggests that these features lack discriminative power and are less effective. However, upon addressing the IMO, we observed an increase in variance within the visual feature space. This is translated into an enhanced class separability as visually demonstrated in Fig. \ref{fig:fig6} where we show the t-SNE visualization of the original and adapted CLIP visual features.

\paragraph{Measuring the Distance Between Training and Target Data}
We also investigated whether the data samples from Google Open Images closely matched the distributions of the downstream datasets we tested on. We aimed to determine if our adapters were potentially overfitting to datasets that resemble each other rather than effectively addressing the broader IMO issue. We use Proxy-A-Distance (PAD) ~\cite{bendavid_2006_analysis} as a measure of the divergence between these datasets. To compute Proxy-A-Distance we create an SVM classifier that is trained to distinguish between the source domain (Google Open Images) and the target domains (other datasets). The PAD is calculated based on the error of this domain classifier:
\begin{equation}
\text{PAD} =  2 \cdot (1-2 \cdot \epsilon)
\label{'eq:pad_equation'}
\end{equation}
where $\epsilon$ is the domain classifier error. The PAD score falls within the range of 0 to 2 - PAD close to 0 corresponds to a classifier accuracy of 50\% indicating that the domain classifier is unable to distinguish between the source and target domains. Conversely, a PAD value of 2 indicates that the classifier is capable of completely discriminating between the two domains, thus they do not follow the same distribution, achieving 100\% accuracy or equivalently with the error rate $\epsilon$ = 0. After computing PAD we measure the correlation between the average difference in performance of the Tip-Adapter and Tip-Adapter++ to determine if there is any connection between improved performance and the proximity of source and target data distributions. The correlation between the two is 0.14 suggesting that there is a weak relation between them. Surprisingly, EuroSAT which has a very different distribution from the training data exhibits the most substantial performance enhancement following the adaptation. In contrast, ImageNet which has a relatively closer resemblance to the training dataset displays a comparatively smaller performance improvement. We thus conclude that we reduced IMO generalizing to datasets that are relatively different from the training adaptation data. PAD for all the datasets can be found in Tab. \ref{table:table3}. 

\hspace{1mm}

\section{Conclusions}

This paper examines the relationship between performance and the intra-modal overlap in training-free methods demonstrating a positive relation between the reduction in intra-modal overlap and improved performance. We show that it's possible to directly correct it within the image space, as opposed to using image-text space as a bridge, by introducing bottleneck adapters to the CLIP vision encoder fine-tuned on a subset from the Google Open Images dataset. We further show that such fine-tuning can be done in both a supervised and self-supervised manner. The supervised intra-modal overlap correction improved the performance by 1.38\% across all the datasets. 

\paragraph{Acknowledgements}
The authors gratefully acknowledge Microsoft's support in providing GPU compute resources through the Microsoft's Accelerating Foundation Models Research grant. We'd also like to acknowledge the support from the University of Bath for studentship.
 
\bibliography{egbib}

\clearpage
\newpage

\appendix
% \onecolumn

\section{Implementation Details}
We select images containing only one labelled class as images from Google Open Images dataset are dense and we also filter some classes that are too general such as "person, people" or that often include other classes such as "hat, shoes" as they appear in dense images. Eventually, we select 2368 classes and 167.287 total images (ablations with more and fewer images shown in the Appendix \ref{sec:apppendix_ablations}) and train the adapters for 1 epoch with a learning rate of 5e-3. We report the average accuracy across 3 different random seeds and perform 10 random augmentations for each training sample. For the unsupervised training we use the same images but train for 10 epochs with learning rate of 5e-5 and momentum teacher of 0.9998. Other hyperparameters are default ones from the official DINO implementation \cite{caron_2021_emerging}. The backbone used in both settings is ViT-B/16, which is compatible with the bottleneck adapter. We used the adapter with the bottleneck of size 64 which achieved the best performance on classification tasks in the original paper. 

\section{Performance Across Fine-grained Datasets}

\begin{figure*}[h!]
\centering
\footnotesize
\includegraphics[width=1\textwidth]{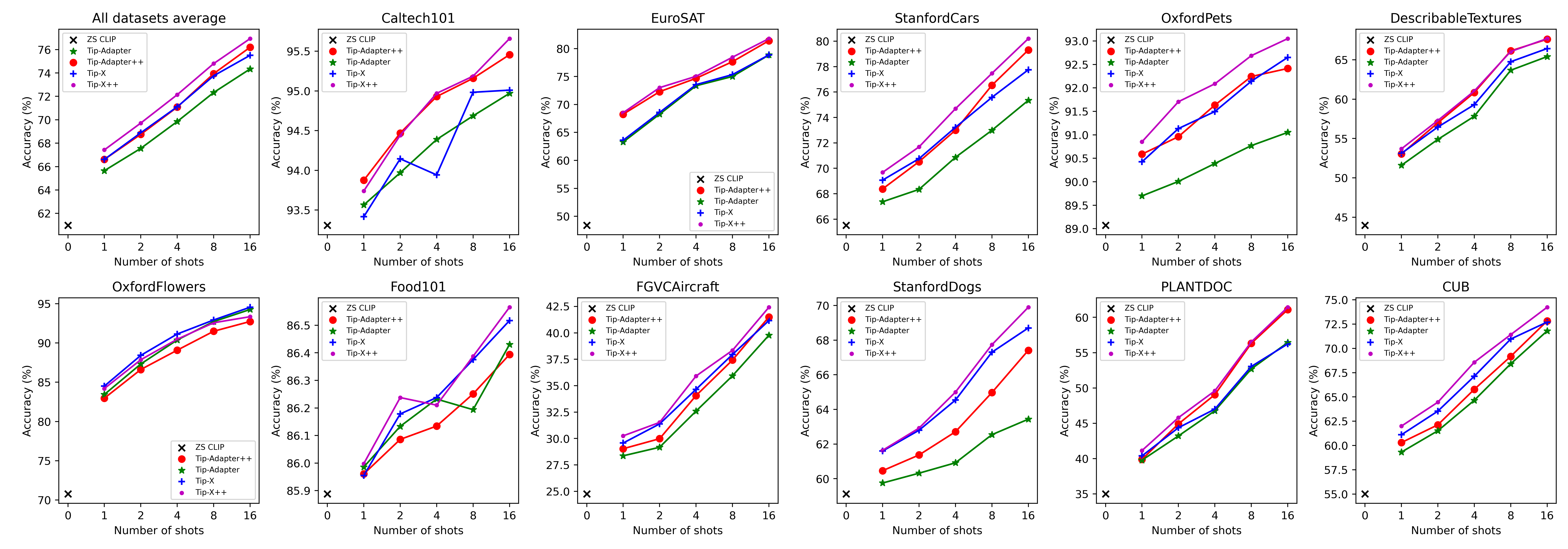}
\vspace*{2mm}
\caption{Performance comparison on 11 fine-grained datasets. Tip-Adapter++ consistently outperforms Tip-Adapter on 9 out of 11 fine-grained datasets with 1 dataset (Food101) achieving similar results and Tip-X++ consistently outperforms Tip-X on 10 out of 11 fine-grained datasets. }
\label{fig:fig4}
\end{figure*}

\section{Justification for few-shot CLIP learning}
In \cite{udandarao2024zeroshot} authors questioned the zero-shot generalization of multimodal models as classes and datasets used to test such capabilities could already be seen in the pretraining set. However, they did identify classes in the long tail of the distribution, where zero-shot performance was notably low, indicating that these classes were either rarely encountered or completely absent during pre-training. We argue that there is therefore still a case to improve the performance for such classes.
We note that few-shot learning is valid especially where the difference between zero-shot and few-shot performance is significant, meaning that classes of those datasets are long tail. For instance, EuroSAT demonstrates low zero-shot performance, but training-free few-shot learning leads to a substantial boost in accuracy of over 23\%. Conversely, certain datasets such as Food101 already exhibit high zero-shot performance, with training-free few-shot learning resulting in only a marginal increase in accuracy of 0.5\%. We improve upon existing training-free few-shot learning methods testing on a variety of datasets including both of these types.

% \newpage
\section{Intra-modal Overlap for All Datasets}
\label{appendix:imo_allds}

\begin{figure}[!th]
\centering\footnotesize
\includegraphics[width=1\textwidth]{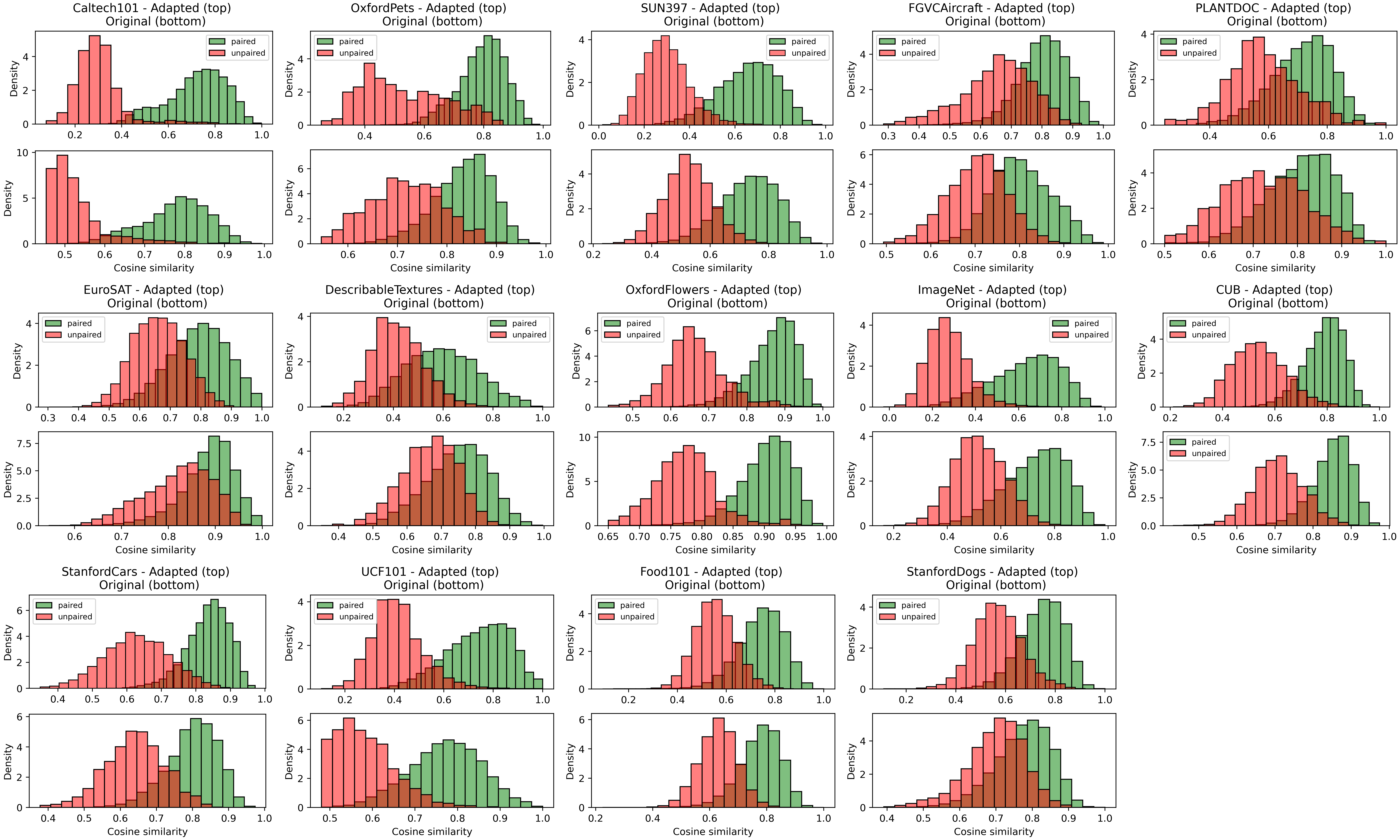}
\fontsize{7}{8}\selectfont
\begin{tabular}{@{}l ccc@{}}
            \toprule 
            Dataset & Adapted Intersection Area (A) & Original Intersection Area (O) & $\Delta$ (O, A)\\
            \midrule 
            Caltech101 & 0.127 & 0.108 & -0.019 \\
            EuroSAT & 0.482 & 0.6 & 0.119 \\
            StanfordCars & 0.215 & 0.323 & 0.108 \\
            OxfordPets & 0.358 & 0.386 & 0.028 \\
            DescribableTextures & 0.47 & 0.633 & 0.163 \\
            UCF101 & 0.187 & 0.219 & 0.033 \\
            SUN397 & 0.146 & 0.26 & 0.114 \\
            OxfordFlowers & 0.168 & 0.158 & -0.01 \\
            Food101 & 0.282 & 0.295 & 0.013 \\
            FGVCAircraft & 0.434 & 0.473 & 0.039 \\
            ImageNet & 0.184 & 0.328 & 0.144 \\
            StanfordDogs & 0.338 & 0.621 & 0.283 \\
            PLANTDOC & 0.514 & 0.61 & 0.096 \\
            CUB & 0.19 & 0.243 & 0.052 \\
            
            \bottomrule

% \caption{Here}
\end{tabular}
\vspace{2mm}
\caption{All datasets intra-modal overlap.}
\label{fig:fig10}
\end{figure}

In Fig. \ref{fig:fig2}
we showed the intra-modal overlap (IMO) measured as an intersection area between cosine similarity distributions of paired and unpaired images for 4 datasets. In Fig. \ref{fig:fig10} we show the same for the remaining datasets, including the not fine-grained ones. The adaptation improves the IMO across 12 out of 14 datasets.

\newpage
\section{Ablations}
\label{sec:apppendix_ablations}
\paragraph{Other Datasets}

We conducted an ablation study across other standard datasets - Cifar100 and PascalVOC. Both of these datasets are of lower quality and less diverse compared to Google Open Images. Consequently, they were unable to decrease intra-modal overlap and improve accuracy to the same extent of Google Open Images when trained in a supervised way.

\begin{table}[!h]
  \centering
  % \footnotesize % Reduce font size
  \fontsize{10}{11}\selectfont
  \setlength{\tabcolsep}{2pt} % Reduce cell padding
  \begin{tabular}{lccc}
    \toprule 
    \textbf{Training Dataset} & \textbf{Avg. IMO} & \textbf{Avg. $\Delta (TA++, TA)$}\\
    \toprule 
    Google Open Images & \textbf{0.083} & \textbf{1.188} \\
    Cifar100 & 0.05 & 0.41 \\
    PascalVOC & 0.01 & 0.12 \\
    \bottomrule
    \end{tabular}
\vspace{2mm}
\caption{Aggregated performance and intra-modal overlap across all datasets and shots for Cifar100, PascalVoc and Google Open Images datasets trained in a supervised way.}
\label{table:mod_perf_datasets}
\end{table}

\paragraph{Number of Samples Sensitivity}
In this analysis, we evaluate the impact of varying the number of samples from the Google Open Images dataset on performance and intra-modal overlap. We observed that an insufficient amount of data (80k samples) did not lead to significant performance improvement while increasing the dataset size to 200k samples did not yield much improvement compared to the 160k samples selected in our main experiments.

\begin{table}[!h]
  \centering
  % \footnotesize % Reduce font size
  \fontsize{10}{11}\selectfont
  \setlength{\tabcolsep}{2pt} % Reduce cell padding
  \begin{tabular}{lccc}
    \toprule 
    \textbf{Number of samples} & \textbf{Avg. IMO} & \textbf{Avg. $\Delta (TA++, TA)$}\\
    \toprule 
    80k & 0.059 & 0.5 \\
    160k & \textbf{0.083} & \textbf{1.188} \\
    200k & 0.076 & 0.82 \\
    \bottomrule
    \end{tabular}
\vspace{2mm}
\caption{Aggregated performance and intra-modal overlap across all datasets and shots for different number of samples from Google Open Images trained in a supervised way.}
\label{table:mod_perf_datasets_samples}
\end{table}

\newpage

\section{Granular Results \& Performance with IMO Relation Across All Datasets}
\label{appendix:b}
\begin{figure}[!h]
\centering\footnotesize
\includegraphics[width=0.5\linewidth]{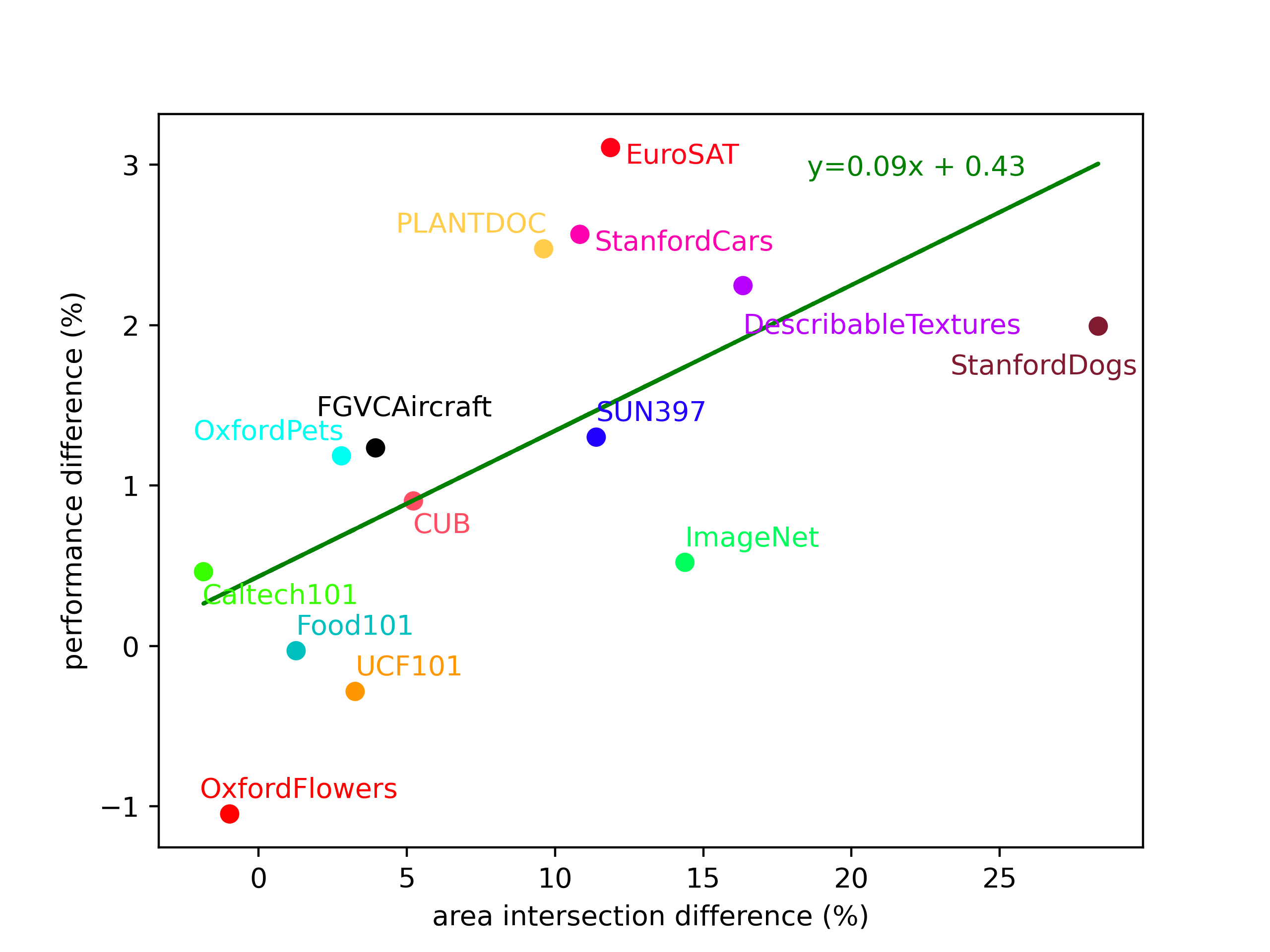}
\vspace{2mm}
\caption{Relation between area intersection difference (intra-modal overlap reduction) between the original and adapted visual encoders vs average performance difference between Tip-Adapter++ and Tip-Adapter with supervised adaptation for all datasets. }
\label{fig:fig12}
\end{figure}

\paragraph{Intra-modal Overlap and Performance Relation}
When we include the not fine-grained datasets as observed in Fig. \ref{fig:fig12} 
the relation between intra-modal overlap reduction and performance improvement stays the same as for only the fine-grained ones reported in Fig. \ref{fig:fig3} in the main paper. 

\newpage
\clearpage

\begin{table}[!ht]
    \centering
    % \fontsize{10pt}{10pt}
    % \footnotesize % Reduce font size
    \fontsize{12.5}{11.5}\selectfont
    \setlength{\tabcolsep}{2pt} % Reduce cell padding
    \resizebox{\textwidth}{!}{
    \begin{tabular}{@{}lcc|cc|cc|ccc@{}} % Adjusted number of columns
    \toprule 
    \textbf{Dataset} & \textbf{Shots} & \textbf{Zero-Shot} & \makecell[c]{\textbf{Tip-Adapter}\\ (TA)} & \makecell[c]{\textbf{Tip-Adapter++}\\ (TA++)} & \makecell[c]{\textbf{Tip-X}\\ (TX)} & \makecell[c]{\textbf{Tip-X}\\ (TX++)} & \makecell[c]{$\Delta$ (TA++, TA)} & \makecell[c]{$\Delta$ (TX++, TX)} & \makecell[c]{$\Delta$ (TA++,TX)}\\
    \midrule 
    % \fontsize{1pt}{1pt}
EuroSAT & 1 & 48.383 & 63.288 & \textbf{68.259} & 63.597 & \textbf{68.527}  & 4.971 & 4.93 & 4.663  \\
EuroSAT & 2 & 48.383 & 68.267 & \textbf{72.292} & 68.576 & \textbf{73.012}  & 4.025 & 4.436 & 3.716  \\
EuroSAT & 4 & 48.383 & 73.354 & \textbf{74.683} & 73.547 & \textbf{75.041}  & 1.329 & 1.494 & 1.136  \\
EuroSAT & 8 & 48.383 & 75.008 & \textbf{77.658} & 75.342 & \textbf{78.457}  & 2.65 & 3.115 & 2.317  \\
EuroSAT & 16 & 48.383 & 78.852 & \textbf{81.407} & 78.864 & \textbf{81.782}  & 2.556 & 2.918 & 2.543  \\
StanfordCars & 1 & 65.514 & 67.367 & \textbf{68.379} & 69.071 & \textbf{69.68}  & 1.011 & 0.609 & -0.692  \\
StanfordCars & 2 & 65.514 & 68.341 & \textbf{70.522} & 70.758 & \textbf{71.683}  & 2.18 & 0.924 & -0.236  \\
StanfordCars & 4 & 65.514 & 70.862 & \textbf{72.997} & 73.221 & \textbf{74.688}  & 2.135 & 1.467 & -0.224  \\
StanfordCars & 8 & 65.514 & 72.988 & \textbf{76.529} & 75.579 & \textbf{77.465}  & 3.54 & 1.886 & 0.949  \\
StanfordCars & 16 & 65.514 & 75.347 & \textbf{79.306} & 77.752 & \textbf{80.201}  & 3.959 & 2.45 & 1.555  \\
PLANTDOC & 1 & 34.994 & 39.78 & \textbf{39.888} & 40.384 & \textbf{41.138}  & 0.108 & 0.755 & -0.496  \\
PLANTDOC & 2 & 34.994 & 43.208 & \textbf{44.912} & 44.373 & \textbf{45.796}  & 1.703 & 1.423 & 0.539  \\
PLANTDOC & 4 & 34.994 & 46.766 & \textbf{49.051} & 47.003 & \textbf{49.59}  & 2.285 & 2.587 & 2.048  \\
PLANTDOC & 8 & 34.994 & 52.695 & \textbf{56.317} & 53.04 & \textbf{56.511}  & 3.622 & 3.471 & 3.277  \\
PLANTDOC & 16 & 34.994 & 56.425 & \textbf{61.082} & 56.231 & \textbf{61.427}  & 4.657 & 5.196 & 4.851  \\
DescribableTextures & 1 & 43.972 & 51.596 & \textbf{53.034} & 53.113 & \textbf{53.684}  & 1.438 & 0.571 & -0.079  \\
DescribableTextures & 2 & 43.972 & 54.886 & \textbf{56.994} & 56.462 & \textbf{57.289}  & 2.108 & 0.827 & 0.532  \\
DescribableTextures & 4 & 43.972 & 57.821 & \textbf{60.835} & 59.299 & \textbf{61.032}  & 3.014 & 1.734 & 1.537  \\
DescribableTextures & 8 & 43.972 & 63.672 & \textbf{66.135} & 64.756 & \textbf{66.056}  & 2.463 & 1.3 & 1.379  \\
DescribableTextures & 16 & 43.972 & 65.406 & \textbf{67.612} & 66.43 & \textbf{67.691}  & 2.206 & 1.261 & 1.182  \\
StanfordDogs & 1 & 59.117 & 59.749 & \textbf{60.461} & 61.596 & \textbf{61.636}  & 0.712 & 0.04 & -1.136  \\
StanfordDogs & 2 & 59.117 & 60.317 & \textbf{61.368} & 62.796 & \textbf{62.92}  & 1.052 & 0.124 & -1.428  \\
StanfordDogs & 4 & 59.117 & 60.917 & \textbf{62.708} & 64.539 & \textbf{64.999}  & 1.791 & 0.46 & -1.831  \\
StanfordDogs & 8 & 59.117 & 62.54 & \textbf{64.971} & 67.302 & \textbf{67.734}  & 2.431 & 0.432 & -2.331  \\
StanfordDogs & 16 & 59.117 & 63.436 & \textbf{67.414} & 68.706 & \textbf{69.902}  & 3.979 & 1.196 & -1.292  \\
SUN397 & 1 & 62.579 & 65.529 & \textbf{66.713} & 66.584 & \textbf{67.058}  & 1.184 & 0.474 & 0.128  \\
SUN397 & 2 & 62.579 & 67.332 & \textbf{68.516} & 68.37 & \textbf{69.093}  & 1.184 & 0.723 & 0.146  \\
SUN397 & 4 & 62.579 & 68.791 & \textbf{70.35} & 70.025 & \textbf{70.929}  & 1.559 & 0.904 & 0.325  \\
SUN397 & 8 & 62.579 & 70.441 & \textbf{71.781} & 71.753 & \textbf{72.809}  & 1.34 & 1.055 & 0.028  \\
SUN397 & 16 & 62.579 & 71.635 & \textbf{72.874} & 72.955 & \textbf{73.776}  & 1.239 & 0.821 & -0.081  \\
FGVCAircraft & 1 & 24.752 & 28.363 & \textbf{29.033} & 29.573 & \textbf{30.253}  & 0.67 & 0.68 & -0.54  \\
FGVCAircraft & 2 & 24.752 & 29.173 & \textbf{29.983} & 31.383 & \textbf{31.523}  & 0.81 & 0.14 & -1.4  \\
FGVCAircraft & 4 & 24.752 & 32.593 & \textbf{34.063} & 34.653 & \textbf{35.914}  & 1.47 & 1.26 & -0.59  \\
FGVCAircraft & 8 & 24.752 & 35.934 & \textbf{37.424} & 37.954 & \textbf{38.344}  & 1.49 & 0.39 & -0.53  \\
FGVCAircraft & 16 & 24.752 & 39.774 & \textbf{41.504} & 41.164 & \textbf{42.424}  & 1.73 & 1.26 & 0.34  \\
OxfordPets & 1 & 89.071 & 89.697 & \textbf{90.588} & 90.424 & \textbf{90.851}  & 0.89 & 0.427 & 0.164  \\
OxfordPets & 2 & 89.071 & 90.006 & \textbf{90.96} & 91.133 & \textbf{91.705}  & 0.954 & 0.572 & -0.173  \\
OxfordPets & 4 & 89.071 & 90.388 & \textbf{91.633} & 91.496 & \textbf{92.087}  & 1.245 & 0.591 & 0.136  \\
OxfordPets & 8 & 89.071 & 90.77 & \textbf{92.241} & 92.141 & \textbf{92.686}  & 1.472 & 0.545 & 0.1  \\
OxfordPets & 16 & 89.071 & 91.051 & \textbf{92.414} & 92.65 & \textbf{93.05}  & 1.363 & 0.4 & -0.236  \\
CUB & 1 & 55.009 & 59.318 & \textbf{60.301} & 61.103 & \textbf{61.995}  & 0.983 & 0.892 & -0.802  \\
CUB & 2 & 55.009 & 61.514 & \textbf{62.128} & 63.536 & \textbf{64.457}  & 0.614 & 0.92 & -1.408  \\
CUB & 4 & 55.009 & 64.652 & \textbf{65.781} & 67.127 & \textbf{68.57}  & 1.129 & 1.443 & -1.346  \\
CUB & 8 & 55.009 & 68.41 & \textbf{69.177} & 70.961 & \textbf{71.415}  & 0.767 & 0.453 & -1.785  \\
CUB & 16 & 55.009 & 71.798 & \textbf{72.823} & 72.711 & \textbf{74.238}  & 1.025 & 1.527 & 0.112  \\
ImageNet & 1 & 68.804 & 69.28 & \textbf{69.536} & 69.389 & \textbf{69.568}  & 0.256 & 0.179 & 0.147  \\
ImageNet & 2 & 68.804 & 69.477 & \textbf{69.805} & 69.509 & \textbf{69.812}  & 0.328 & 0.303 & 0.297  \\
ImageNet & 4 & 68.804 & 69.791 & \textbf{70.359} & 69.864 & \textbf{70.359}  & 0.569 & 0.495 & 0.495  \\
ImageNet & 8 & 68.804 & 70.249 & \textbf{70.949} & 70.459 & \textbf{71.012}  & 0.699 & 0.553 & 0.489  \\
ImageNet & 16 & 68.804 & 70.753 & \textbf{71.505} & 70.973 & \textbf{71.587}  & 0.753 & 0.613 & 0.532  \\
Caltech101 & 1 & 93.306 & 93.563 & \textbf{93.874} & 93.414 & \textbf{93.739}  & 0.311 & 0.325 & 0.46  \\
Caltech101 & 2 & 93.306 & 93.969 & \textbf{94.469} & 94.145 & \textbf{94.442}  & 0.5 & 0.297 & 0.325  \\
Caltech101 & 4 & 93.306 & 94.388 & \textbf{94.929} & 93.942 & \textbf{94.97}  & 0.541 & 1.028 & 0.987  \\
Caltech101 & 8 & 93.306 & 94.686 & \textbf{95.159} & 94.983 & \textbf{95.186}  & 0.473 & 0.203 & 0.176  \\
Caltech101 & 16 & 93.306 & 94.97 & \textbf{95.456} & 95.01 & \textbf{95.659}  & 0.487 & 0.649 & 0.446  \\
Food101 & 1 & 85.888 & \textbf{85.986} & 85.96 & 85.955 & \textbf{85.998}  & -0.025 & 0.043 & 0.006  \\
Food101 & 2 & 85.888 & \textbf{86.133} & 86.086 & 86.178 & \textbf{86.238}  & -0.047 & 0.059 & -0.092  \\
Food101 & 4 & 85.888 & \textbf{86.232} & 86.134 & \textbf{86.238} & 86.21  & -0.098 & -0.028 & -0.103  \\
Food101 & 8 & 85.888 & 86.194 & \textbf{86.251} & 86.375 & \textbf{86.387}  & 0.057 & 0.012 & -0.124  \\
Food101 & 16 & 85.888 & \textbf{86.43} & 86.394 & 86.517 & \textbf{86.565}  & -0.036 & 0.048 & -0.123  \\
UCF101 & 1 & 67.46 & 71.716 & \textbf{72.024} & 72.553 & \textbf{72.667}  & 0.308 & 0.115 & -0.529  \\
UCF101 & 2 & 67.46 & 73.777 & \textbf{73.857} & 75.17 & \textbf{75.24}  & 0.079 & 0.07 & -1.313  \\
UCF101 & 4 & 67.46 & \textbf{74.007} & 73.795 & \textbf{75.399} & 75.17  & -0.211 & -0.229 & -1.604  \\
UCF101 & 8 & 67.46 & \textbf{77.284} & 76.509 & 78.298 & \textbf{78.377}  & -0.775 & 0.079 & -1.789  \\
UCF101 & 16 & 67.46 & \textbf{78.421} & 77.602 & 78.773 & \textbf{79.038}  & -0.819 & 0.264 & -1.172  \\
OxfordFlowers & 1 & 70.767 & \textbf{83.435} & 82.961 & \textbf{84.504} & 84.193  & -0.474 & -0.311 & -1.543  \\
OxfordFlowers & 2 & 70.767 & \textbf{87.319} & 86.615 & \textbf{88.415} & 87.86  & -0.704 & -0.555 & -1.8  \\
OxfordFlowers & 4 & 70.767 & \textbf{90.378} & 89.078 & \textbf{91.135} & 90.472  & -1.299 & -0.663 & -2.057  \\
OxfordFlowers & 8 & 70.767 & \textbf{92.719} & 91.487 & \textbf{92.922} & 92.57  & -1.232 & -0.352 & -1.435  \\
OxfordFlowers & 16 & 70.767 & \textbf{94.262} & 92.732 & \textbf{94.546} & 93.341  & -1.529 & -1.204 & -1.814  \\

\midrule
Average fine-grained & 1 & 60.979 & 65.649 & \textbf{66.613} & 66.612 & \textbf{67.427}  & 0.963 & 0.815 & 0.0  \\
Average fine-grained & 2 & 60.979 & 67.558 & \textbf{68.757} & 68.887 & \textbf{69.72}  & 1.2 & 0.834 & -0.13  \\
Average fine-grained & 4 & 60.979 & 69.85 & \textbf{71.081} & 71.109 & \textbf{72.143}  & 1.231 & 1.034 & -0.028  \\
Average fine-grained & 8 & 60.979 & 72.329 & \textbf{73.941} & 73.76 & \textbf{74.801}  & 1.612 & 1.041 & 0.181  \\
Average fine-grained & 16 & 60.979 & 74.341 & \textbf{76.195} & 75.507 & \textbf{76.935}  & 1.854 & 1.427 & 0.688  \\
Average all & 1 & 62.115 & 66.333 & \textbf{67.215} & 67.233 & \textbf{67.928}  & 0.882 & 0.695 & -0.018  \\
Average all & 2 & 62.115 & 68.123 & \textbf{69.179} & 69.343 & \textbf{70.076}  & 1.056 & 0.733 & -0.164  \\
Average all & 4 & 62.115 & 70.067 & \textbf{71.171} & 71.249 & \textbf{72.145}  & 1.104 & 0.896 & -0.078  \\
Average all & 8 & 62.115 & 72.399 & \textbf{73.756} & 73.705 & \textbf{74.644}  & 1.357 & 0.939 & 0.052  \\
Average all & 16 & 62.115 & 74.183 & \textbf{75.723} & 75.235 & \textbf{76.477}  & 1.541 & 1.243 & 0.489  \\
    \bottomrule
    \end{tabular}}
    \vspace{2mm}
    \caption{Average results by number of shots over 3 seeds.}
    \label{table:table2}
\end{table}

\newpage
\clearpage

% \newpage
% \clearpage

% \onecolumn
\section{Unsupervised Training}
\label{appendix:c}

\begin{figure}[!h]
\centering\footnotesize
\includegraphics[width=0.5\linewidth]{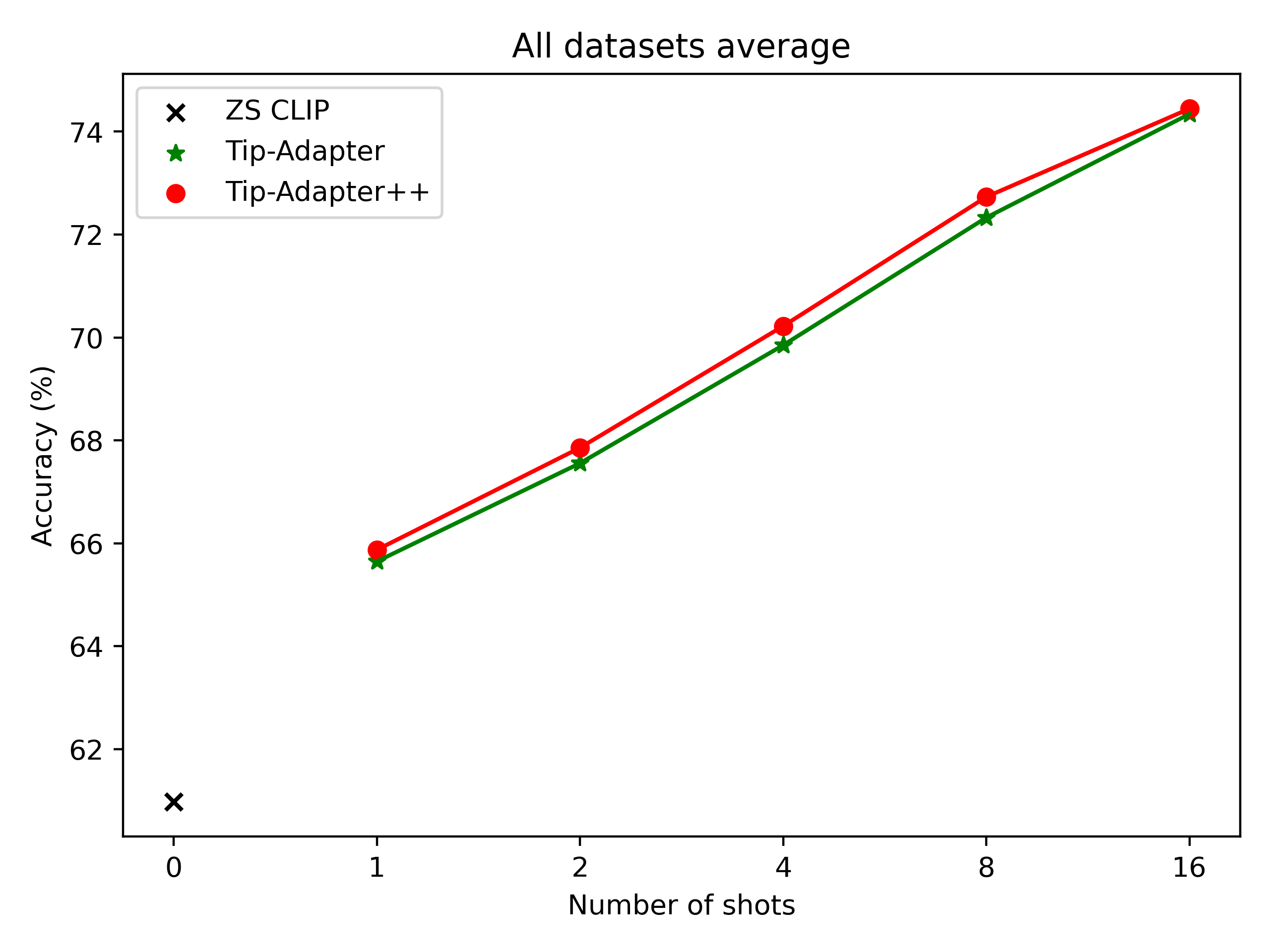}
\caption{Performance unsupervised intra-modal overlap correction. Figure shows the average performance of Tip-Adapter and Tip-Adapter++ across different shots for fine-grained datasets. }
\label{fig:fig11}
\end{figure}

\paragraph{Results}
In Fig. \ref{fig:fig11} and Table \ref{table:table6} we compare the performance of Tip-Adapter and Tip-Adapter++ (similar results for Tip-X vs Tip-X++ that we omit) observing that with unsupervised adaptation Tip-Adapter++ outperforms Tip-Adapter on 7 out of 14 datasets. These results are worse than the supervised counterpart, however, we believe that it is interesting to correct the intra-modal overlap through adaptation training adapters in an unsupervised way. As future work we will try to do it with a bigger and more diverse dataset.  

\begin{table}[!th]
\centering
% \footnotesize % Reduce font size
\setlength{\tabcolsep}{2pt} % Reduce cell padding
\fontsize{9}{9}\selectfont
\resizebox{0.8\textwidth}{!}{
\begin{tabular}{@{}lcccc@{}}
            \toprule 
            \textbf{Dataset} & \textbf{Zero-Shot} & \makecell[c]{\textbf{Tip-Adapter}\\ (TA)} & \makecell[c]{\textbf{Tip-Adapter++}\\ (TA++)} & \makecell[c]{$\Delta$ (TA++, TA)} \\
            \midrule 
EuroSAT & 48.383 & 71.754 & \textbf{74.915} & 3.161 \\
DescribableTextures & 43.972 & 58.676 & \textbf{59.18} & 0.504 \\
SUN397 & 62.579 & 68.783 & \textbf{69.115} & 0.332 \\
StanfordCars & 65.514 & 70.981 & \textbf{71.283} & 0.302 \\
UCF101 & 67.46 & 75.041 & \textbf{75.286} & 0.245 \\
OxfordFlowers & 70.767 & 89.622 & \textbf{89.712} & 0.089 \\
OxfordPets & 89.071 & 90.382 & \textbf{90.464} & 0.082 \\
Food101 & 85.888 & \textbf{86.195} & 86.182 & -0.013 \\
ImageNet & 68.801 & \textbf{69.911} & 69.897 & -0.014 \\
PLANTDOC & 34.994 & \textbf{47.775} & 47.749 & -0.026 \\
FGVCAircraft & 24.752 & \textbf{33.167} & 33.071 & -0.096 \\
Caltech101 & 93.306 & \textbf{94.315} & 94.191 & -0.124 \\
StanfordDogs & 59.117 & \textbf{61.392} & 61.242 & -0.15 \\
CUB & 55.009 & \textbf{65.138} & 64.494 & -0.644 \\
\midrule
Average fine-grained & 60.979 & 69.945 & \textbf{70.226} & 0.281 \\
Average all & 62.115 & 70.224 & \textbf{70.484} & 0.261 \\
            \bottomrule
\end{tabular}}
\vspace{2mm}
\caption{Performance unsupervised intra-modal overlap correction. Table shows the comparison between average performance of Tip-Adapter and Tip-Adapter++ across different shots for all the datasets.}
\label{table:table6}

\end{table}

\begin{figure}[!htb]
\begin{minipage}{.5\textwidth}
  \centering
  \includegraphics[width=\linewidth]{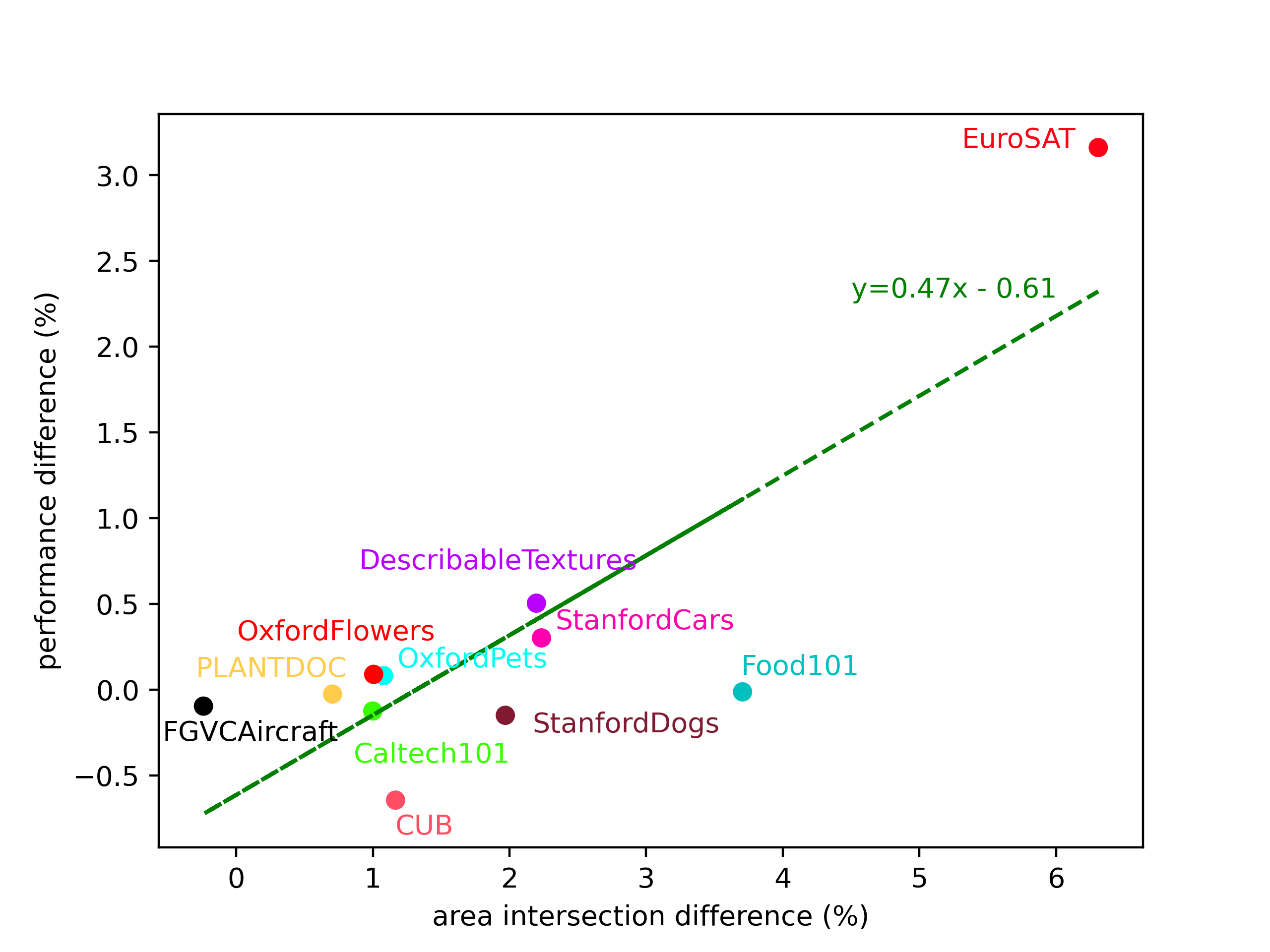}
  \vspace{1mm}
  \caption*{(a) Fine-grained datasets}
\end{minipage}
\begin{minipage}{.5\textwidth}
  \centering
  \includegraphics[width=\linewidth]{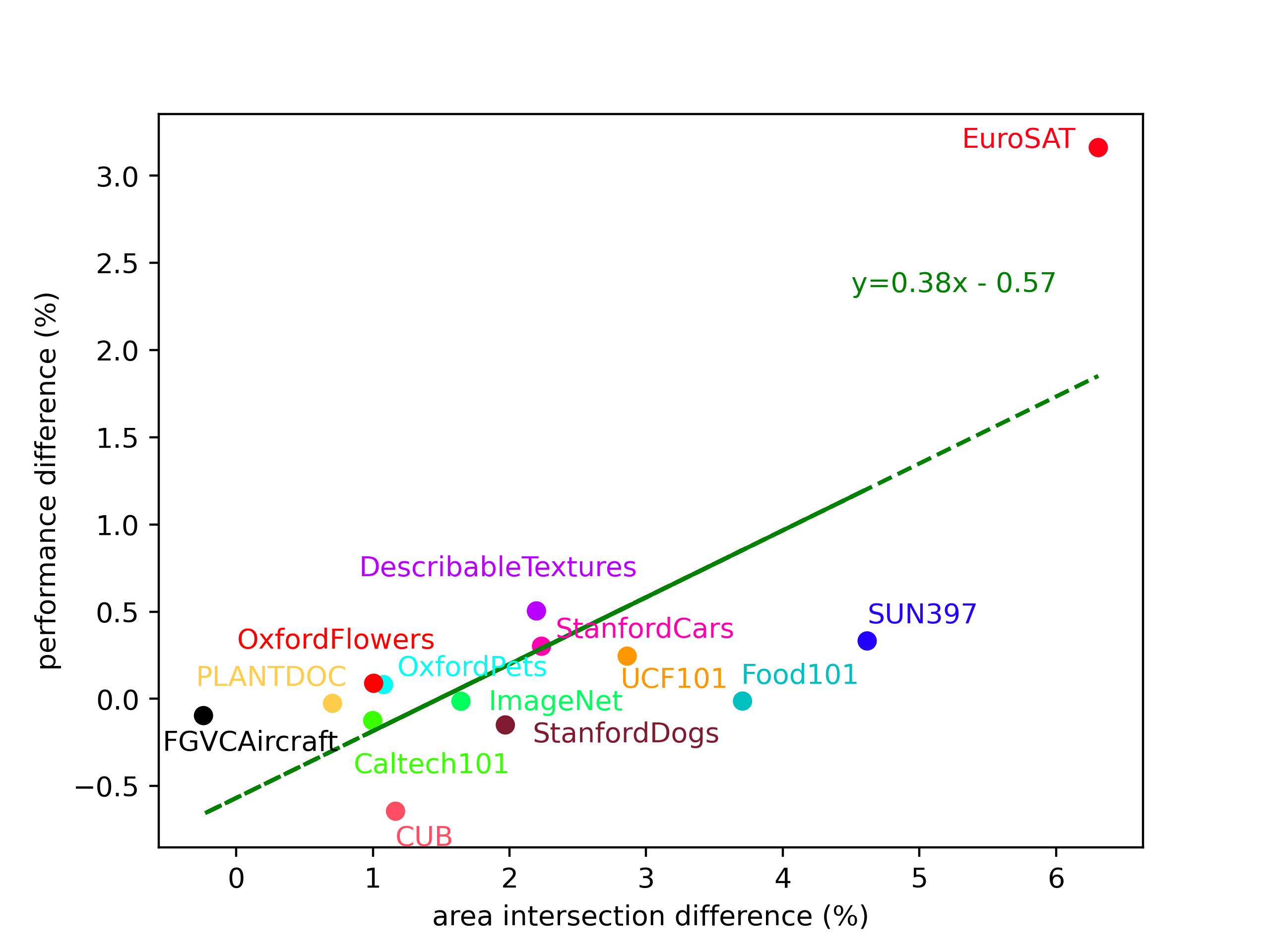}
  \vspace{1mm}
  \caption*{(b) All datasets}
\end{minipage}
\vspace*{5mm}
\caption{Relation between area intersection difference (intra-modal overlap reduction) between the original and adapted visual encoders vs average performance difference between Tip-Adapter++ and Tip-Adapter with unsupervised adaptation. Fig. (a) shows this relation for fine-grained datasets while Fig. (b) for all the datasets. }
\label{fig:fig9}
\end{figure}

\newpage
\paragraph{Performance and the Relation with Intra-modal Overlap of Unsupervised Adaptation}
In Fig. \ref{fig:fig9} we observe a positive relation between the difference in intersection area and the average performance difference, mirroring the pattern seen in the supervised counterpart.

\onecolumn
\section{LoRA Adapter}
\label{appendix:e}

We perform an ablation study implementing the LoRA \cite{hu_2021_lora} adapter rather than the bottleneck adapter \cite{chen_2022_adaptformer}. LoRA adapter is applied to the self-attention at each layer of the visual encoder. The results presented in Table \ref{table:table7} indicate a significant degradation in performance compared to using the bottleneck adapter. We attribute the inferior performance of LoRA to the fact that the bottleneck adapter keeps the CLIP visual encoder weights frozen, maintaining extensive knowledge about different classes acquired during CLIP pretraining and only slightly adjusts the features with the effect of reducing the intra-modal overlap, while the application of LoRA adapters breaks that knowledge leading to inferior performance. 

\begin{table}[!tbh]
\centering
\footnotesize % Reduce font size
\setlength{\tabcolsep}{2pt} % Reduce cell padding
\resizebox{0.8\textwidth}{!}{
\begin{tabular}{@{}lcccc@{}}
            \toprule 
            \textbf{Dataset} & \textbf{Zero-Shot} & \makecell[c]{\textbf{Tip-Adapter}\\ (TA)} & \makecell[c]{\textbf{Tip-Adapter++}\\ (TA++)} & \makecell[c]{$\Delta$ (TA++, TA)} \\
            \midrule 
            OxfordPets & 89.071 & 90.382 & 89.97 & -0.412 \\
            Food101 & 85.888 & 86.195 & 85.984 & -0.211 \\
            Caltech101 & 93.306 & 94.315 & 93.915 & -0.4 \\
            StanfordDogs & 59.117 & 61.392 & 61.156 & -0.235 \\
            ImageNet & 68.804 & 69.911 & 69.374 & -0.537 \\
            SUN397 & 62.579 & 68.783 & 66.516 & -2.267 \\
            UCF101 & 67.46 & 75.041 & 71.478 & -3.563 \\
            EuroSAT & 48.383 & 71.754 & 69.165 & -2.588 \\
            StanfordCars & 65.514 & 70.981 & 67.798 & -3.184 \\
            PLANTDOC & 34.994 & 47.775 & 44.489 & -3.286 \\
            CUB & 55.009 & 65.138 & 58.444 & -6.695 \\
            DescribableTextures & 43.972 & 58.676 & 51.052 & -7.624 \\
            FGVCAircraft & 24.752 & 33.167 & 26.163 & -7.005 \\
            OxfordFlowers & 70.767 & 89.622 & 79.878 & -9.744 \\
            \bottomrule

% \caption{Here}
\end{tabular}}
\vspace{2mm}
\caption{Performance comparison between average performance of Tip-Adapter and Tip-Adapter++ for each dataset across different shots using LoRA Adapter.}
\label{table:table7}
\end{table}

\onecolumn
\section{APE Training-free Method}
\label{appendix:d}

\paragraph{Method Description}
APE \cite{zhu_2023_not} is a training-free method where most discriminative features from the last vision and text CLIP layers are selected eliminating less discriminative feature channels based on a prior refinement module. They employ two criteria for this selection: inter-class similarity and variance. Inter-class similarity criterion focuses on extracting feature channels that minimize the inter-class similarity. On the other hand the inter-class variance criterion eliminates feature channels that exhibit minimal variation between categories as these channels have little impact on classification. These two criteria are then combined to extract the most discriminative features. 
With such refined features, indicated by $ ' $ symbol, the authors compute APE classification logits for a test image. These are given by the sum of CLIP zero-shot logits and Tip-Adapter affinity matrix but weighted by the uncertainty of CLIP logits based on few-shot training examples instead of training labels. To compute these weights, they calculate the Kullback-Leibler (KL) divergence between the zero-shot CLIP classification probabilities derived from training data features $F_{train}$ as defined in Eq. \ref{eq:ta_features} 
and classifier weight matrix $W$ and the true labels $L_{train}$ as defined in Eq. \ref{eq:ta_labels} 
in the main paper: \\
\begin{equation}
    \text{ APEweights } = exp(\gamma D_{KL}(F'_{train} W'^T | L_{train})), \in \mathbb{R}^{1 \times NK}
    \label{eq:APE_weights}
\end{equation} \\
Where $ ' $ indicates that the features were refined with the refinement module and $\gamma$ is a smoothing factor. 

These weights reflect the divergence between the true and zero-shot CLIP predicted labels. For classes where there is more uncertainty in zero-shot CLIP prediction, i.e., where the KL divergence is high, we need to rely more on the cache model and vice versa. Final prediction logits for APE are given by: \\
\begin{equation}
    \text{ APElogits } = \text{CLIPlogits} + \alpha A' (\text{diag(APEweights)} \ L_{train})
    \label{eq:APE_logits}
\end{equation}\\
Where \textit{A'} is the affinity matrix as defined in Eq. \ref{eq:ta_affinity}
but with refined features,\textit{diag} is the diagonalization operator and $\alpha$ is a weighting constant. 

Replacing the affinity matrix $A'$ with the intra-modal overlap corrected one, $Y'$, as in Eq. \ref{eq:ta_affinity_modgap} 
we obtain APE++: \\
\begin{equation}
    \text{ APElogits++ } = \text{CLIPlogits} + \alpha Y' (\text{diag(APEweights)} \ L_{train})
    \label{eq:APE_logits}
\end{equation}

\begin{figure}[!thb]
\centering\footnotesize
% \captionsetup{justification=centering,margin=3.5cm}
\includegraphics[width=0.6\linewidth]{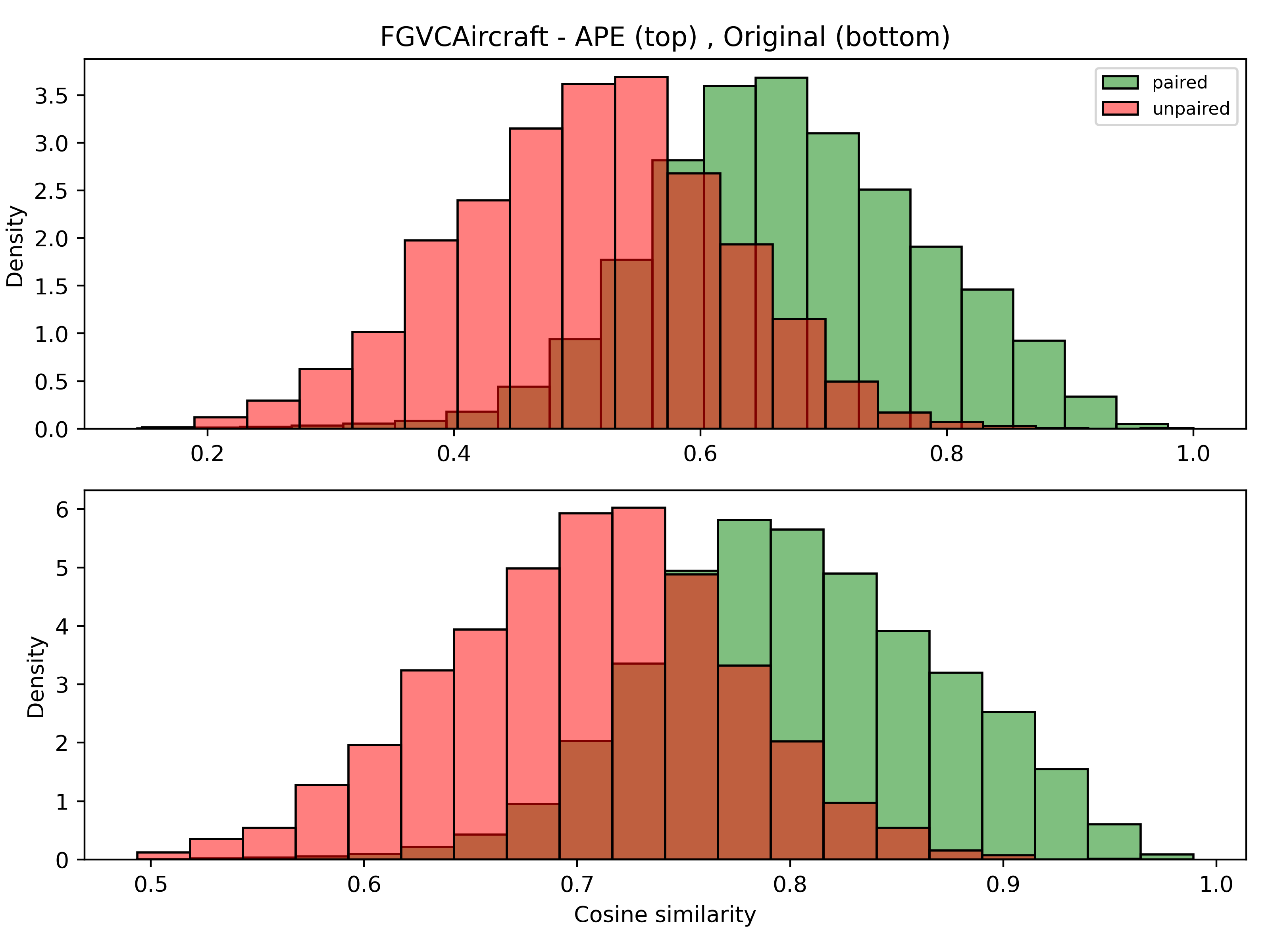}
\caption{Intra and inter-class cosine similarity on FGVCAircraft after APE refinement. Both intra-class and inter-class similarity decreases almost not affecting the intra-modal overlap. }
\label{fig:fig7}
\end{figure}

\paragraph{Intra-modal Overlap After Features Pruning}
As discussed above authors of APE proposed a method to select more discriminative features by eliminating certain feature channels based on inter-class similarity criterion. This has the effect of shifting the unpaired distribution of cosine similarities to the left but, as we illustrate in Fig. \ref{fig:fig7} and in Tab. \ref{table:table5} it also moves the distribution of the paired images to the left thus either changing only slightly or making worse the intra-modal overlap in most cases. 

% \twocolumn
\begin{table}[!tbh]
\centering
\footnotesize % Reduce font size
\setlength{\tabcolsep}{2pt} % Reduce cell padding
\resizebox{\textwidth}{!}{
\begin{tabular}{@{}l ccc@{}}
            \toprule 
            Dataset & APE Intersection Area (APE) & Original Intersection Area (O) & $\Delta$ (O, APE)\\
            \midrule 
            Caltech101 & 0.36 & 0.108 & -0.252 \\
            EuroSAT & 0.61 & 0.6 & -0.01 \\
            StanfordCars & 0.484 & 0.323 & -0.161 \\
            OxfordPets & 0.464 & 0.386 & -0.078 \\
            DescribableTextures & 0.566 & 0.633 & 0.067 \\
            UCF101 & 0.311 & 0.219 & -0.091 \\
            SUN397 & 0.232 & 0.26 & 0.027 \\
            OxfordFlowers & 0.2 & 0.158 & -0.042 \\
            Food101 & 0.26 & 0.295 & 0.035 \\
            FGVCAircraft & 0.4731 & 0.473 & -0.0001 \\
            ImageNet & 0.292 & 0.328 & 0.036 \\
            StanfordDogs & 0.571 & 0.621 & 0.05 \\
            PLANTDOC & 0.644 & 0.61 & -0.034 \\
            CUB & 0.246 & 0.243 & -0.003 \\
            \bottomrule

% \caption{Here}
\end{tabular}}
\vspace{2mm}
\caption{Intra-modal overlap after adaptive features refinement.}
\label{table:table5}
\end{table}

\paragraph{Results}
In Tab. \ref{table:table9} we include the results with APE model for completeness. We can observe that in 10 out of 14 datasets APE++ outperforms APE although the margin of improvement is often smaller compared to the other training-free methods. This observed trend is attributed to the impact of features pruning. Indeed, as shown in Tab. \ref{table:table10} without feature pruning APE++ exhibits a more substantial performance improvement over APE, similar to the enhancements observed with Tip-Adapter and Tip-X. This is interesting as it indicates that by pruning features, while the intra-modal overlap is not reduced (implying the paired and unpaired samples are close), the features do lie on different sides of the decision boundary of the classifier. This would be a reduced sub-space of features that fits the features based on the decision boundary of the classifier. However, such an approach would not necessarily be robust or have the variance properties. We will investigate opportunities for residual subspace learning that are robust and with variance that explore the decision boundary of classifiers in the future. 

\begin{figure}[!hbt]
\centering
\footnotesize
\includegraphics[width=1\textwidth]{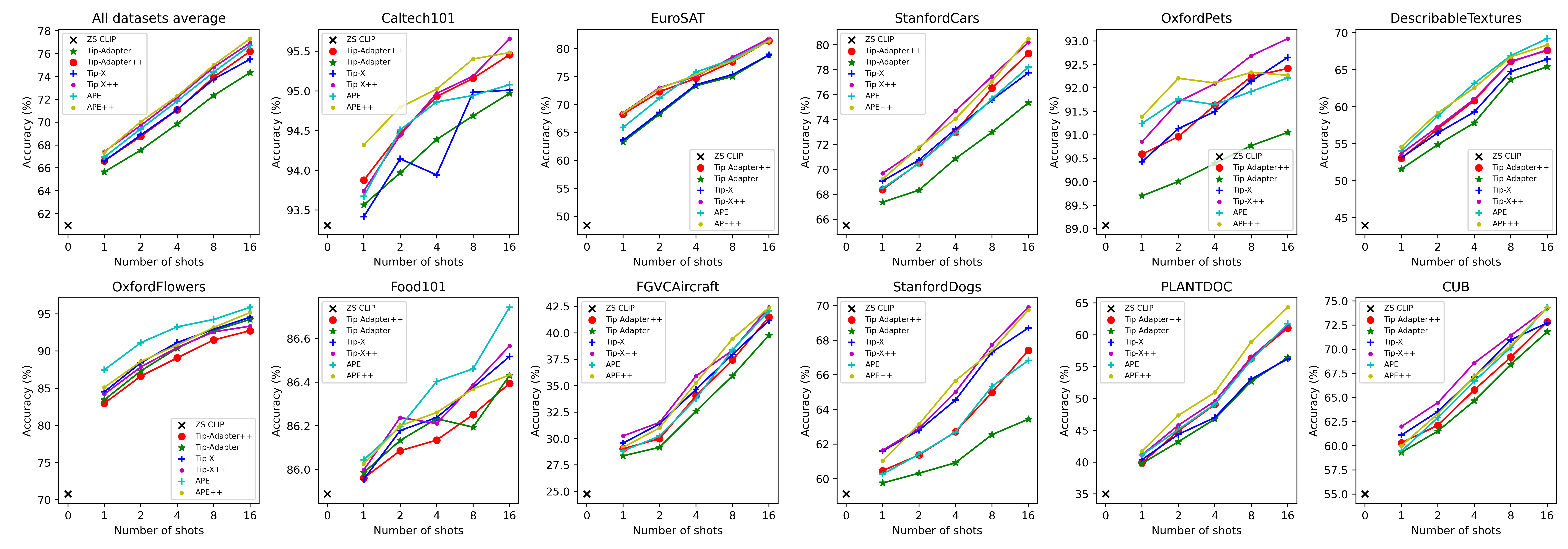}
\vspace{2mm}
\caption{Performance comparison on fine-grained datasets including APE method}
\label{fig:fig8}
\end{figure}

\begin{table}[!tbh]
    \centering
    \fontsize{7}{8.5}\selectfont % Reduce font size
    \setlength{\tabcolsep}{2pt} % Reduce cell padding
    \resizebox{\textwidth}{!}{
    \begin{tabular}{@{}lc|cc|cc|cc|ccccc@{}} % Adjusted number of columns
    \toprule 
    \textbf{Dataset} & \textbf{Zero-Shot} & \makecell[c]{\textbf{Tip-Adapter}\\ (TA)} & \makecell[c]{\textbf{Tip-Adapter++}\\ (TA++)} & \makecell[c]{\textbf{Tip-X}\\ (TX)} & \makecell[c]{\textbf{Tip-X++}\\ (TX++)} & \textbf{APE} & \textbf{APE++} &  \makecell[c]{$\Delta$ (TA++, TA)} & \makecell[c]{$\Delta$ (TX++,TX)} & \makecell[c]{$\Delta$ (TA++, TX)} & $\Delta$(APE++, APE)\\
    \midrule 
    EuroSAT & 48.383 & 71.754 & \textbf{74.86} & 71.985 & \textbf{75.364} & 74.486 &  \textbf{75.165} &  3.106 & 3.379 & 2.875 & 0.679 \\
StanfordCars & 65.514 & 70.981 & \textbf{73.546} & 73.276 & \textbf{74.744} & 73.156 &  \textbf{74.524} &  2.565 & 1.467 & 0.27 & 1.368 \\
PLANTDOC & 34.994 & 47.775 & \textbf{50.25} & 48.206 & \textbf{50.893} & 50.63 &  \textbf{52.652} &  2.475 & 2.687 & 2.044 & 2.022 \\
DescribableTextures & 43.972 & 58.676 & \textbf{60.922} & 60.012 & \textbf{61.151} & \textbf{62.411} &  62.281 &  2.246 & 1.139 & 0.91 & -0.13 \\
StanfordDogs & 59.117 & 61.392 & \textbf{63.385} & 64.988 & \textbf{65.438} & 63.304 &  \textbf{65.39} &  1.993 & 0.45 & -1.603 & 2.086 \\
SUN397 & 62.579 & 68.746 & \textbf{70.047} & 69.938 & \textbf{70.733} & 70.447 &  \textbf{71.016} &  1.301 & 0.795 & 0.109 & 0.569 \\
FGVCAircraft & 24.752 & 33.167 & \textbf{34.401} & 34.945 & \textbf{35.692} & 34.659 &  \textbf{35.454} &  1.234 & 0.746 & -0.544 & 0.795 \\
OxfordPets & 89.071 & 90.382 & \textbf{91.567} & 91.569 & \textbf{92.076} & 91.756 &  \textbf{92.06} &  1.185 & 0.507 & -0.002 & 0.304 \\
CUB & 55.009 & 65.138 & \textbf{66.042} & 67.088 & \textbf{68.135} & 66.709 &  \textbf{67.033} &  0.904 & 1.047 & -1.046 & 0.324 \\
ImageNet & 68.804 & 69.91 & \textbf{70.431} & 70.039 & \textbf{70.468} & 70.29 &  \textbf{70.827} &  0.521 & 0.429 & 0.392 & 0.537 \\
Caltech101 & 93.306 & 94.315 & \textbf{94.778} & 94.299 & \textbf{94.799} & 94.613 &  \textbf{95.005} &  0.462 & 0.5 & 0.479 & 0.392 \\
Food101 & 85.888 & \textbf{86.195} & 86.165 & 86.253 & \textbf{86.28} & \textbf{86.369} &  86.257 &  -0.03 & 0.027 & -0.088 & -0.112 \\
UCF101 & 67.46 & \textbf{75.041} & 74.757 & 76.038 & \textbf{76.098} & \textbf{77.129} &  75.912 &  -0.284 & 0.06 & -1.281 & -1.217 \\
OxfordFlowers & 70.767 & \textbf{89.622} & 88.575 & \textbf{90.305} & 89.687 & \textbf{92.394} &  90.562 &  -1.048 & -0.617 & -1.73 & -1.832 \\
\midrule
Average fine-grained & 60.979 & 69.945 & \textbf{71.317} & 71.175 & \textbf{72.205} & 71.863 &  \textbf{72.398} &  1.372 & 1.03 & 0.142 & 0.535 \\
Average all & 62.115 & 70.221 & \textbf{71.409} & 71.353 & \textbf{72.254} & 72.025 &  \textbf{72.438} &  1.188 & 0.901 & 0.056 & 0.413 \\
    \bottomrule
    \end{tabular}}
    \vspace{2mm}
    \caption{Average performance datasets across all shots including APE.}
    \label{table:table9}
\end{table}

% \newpage

\begin{table}[!tbh]
    \centering
    \fontsize{7}{8.5}\selectfont % Reduce font size
    \setlength{\tabcolsep}{2pt} % Reduce cell padding
    \resizebox{\textwidth}{!}{
    \begin{tabular}{@{}lc|cc|cc|cc|ccccc@{}} % Adjusted number of columns
    \toprule 
    \textbf{Dataset} & \textbf{Zero-Shot} & \makecell[c]{\textbf{Tip-Adapter}\\ (TA)} & \makecell[c]{\textbf{Tip-Adapter++}\\ (TA++)} & \makecell[c]{\textbf{Tip-X}\\ (TX)} & \makecell[c]{\textbf{Tip-X++}\\ (TX++)} & \textbf{APE} & \textbf{APE++} &  \makecell[c]{$\Delta$ (TA++, TA)} & \makecell[c]{$\Delta$ (TX++,TX)} & \makecell[c]{$\Delta$ (TA++, TX)} & $\Delta$(APE++, APE)\\
    \midrule 
    EuroSAT & 48.383 & 71.754 & \textbf{74.86} & 71.985 & \textbf{75.364} & 72.61 &  \textbf{75.677} &  3.106 & 3.379 & 2.875 & 3.067 \\
StanfordCars & 65.514 & 70.981 & \textbf{73.546} & 73.276 & \textbf{74.744} & 71.596 &  \textbf{73.935} &  2.565 & 1.467 & 0.27 & 2.339 \\
PLANTDOC & 34.994 & 47.775 & \textbf{50.25} & 48.206 & \textbf{50.893} & 48.491 &  \textbf{51.397} &  2.475 & 2.687 & 2.044 & 2.906 \\
DescribableTextures & 43.972 & 58.676 & \textbf{60.922} & 60.012 & \textbf{61.151} & 59.421 &  \textbf{61.446} &  2.246 & 1.139 & 0.91 & 2.025 \\
StanfordDogs & 59.117 & 61.392 & \textbf{63.385} & 64.988 & \textbf{65.438} & 61.815 &  \textbf{64.314} &  1.993 & 0.45 & -1.603 & 2.499 \\
SUN397 & 62.579 & 68.746 & \textbf{70.047} & 69.938 & \textbf{70.733} & 69.52 &  \textbf{70.855} &  1.301 & 0.795 & 0.109 & 1.335 \\
FGVCAircraft & 24.752 & 33.167 & \textbf{34.401} & 34.945 & \textbf{35.692} & 33.595 &  \textbf{34.595} &  1.234 & 0.746 & -0.544 & 1.0 \\
OxfordPets & 89.071 & 90.382 & \textbf{91.567} & 91.569 & \textbf{92.076} & 91.102 &  \textbf{91.694} &  1.185 & 0.507 & -0.002 & 0.592 \\
CUB & 55.009 & 65.138 & \textbf{66.042} & 67.088 & \textbf{68.135} & 65.466 &  \textbf{66.46} &  0.904 & 1.047 & -1.046 & 0.994 \\
ImageNet & 68.804 & 69.91 & \textbf{70.431} & 70.039 & \textbf{70.468} & 70.219 &  \textbf{70.827} &  0.521 & 0.429 & 0.392 & 0.608 \\
Caltech101 & 93.306 & 94.315 & \textbf{94.778} & 94.299 & \textbf{94.799} & 94.723 &  \textbf{95.064} &  0.462 & 0.5 & 0.479 & 0.341 \\
Food101 & 85.888 & \textbf{86.195} & 86.165 & 86.253 & \textbf{86.28} & \textbf{86.39} &  86.335 &  -0.03 & 0.027 & -0.088 & -0.055 \\
UCF101 & 67.46 & \textbf{75.041} & 74.757 & 76.038 & \textbf{76.098} & \textbf{75.994} &  75.545 &  -0.284 & 0.06 & -1.281 & -0.449 \\
OxfordFlowers & 70.767 & \textbf{89.622} & 88.575 & \textbf{90.305} & 89.687 & \textbf{90.613} &  89.081 &  -1.048 & -0.617 & -1.73 & -1.532 \\
\midrule
Average fine-grained & 60.979 & 69.945 & \textbf{71.317} & 71.175 & \textbf{72.205} & 70.529 &  \textbf{71.818} &  1.372 & 1.03 & 0.142 & 1.289 \\
Average all & 62.115 & 70.221 & \textbf{71.409} & 71.353 & \textbf{72.254} & 70.825 &  \textbf{71.945} &  1.188 & 0.901 & 0.056 & 1.12 \\

    \bottomrule
    \end{tabular}}
    \vspace{2mm}
    \caption{Average performance datasets across all shots including APE without features pruning.}
    \label{table:table10}
\end{table}

\newpage

\begin{table}[!tbh]
    \centering
    % \fontsize{10pt}{10pt}
    % \footnotesize % Reduce font size
    \fontsize{8}{10.2}\selectfont
    \setlength{\tabcolsep}{2pt} % Reduce cell padding
    \resizebox{\textwidth}{!}{
    \begin{tabular}{@{}lcc|cc|cc|cc|cccc@{}} % Adjusted number of columns
    \toprule 
    \textbf{Dataset} & \textbf{Shots} & \textbf{Zero-Shot} & \makecell[c]{\textbf{Tip-Adapter}\\ (TA)} & \makecell[c]{\textbf{Tip-Adapter++}\\ (TA++)} & \makecell[c]{\textbf{Tip-X}\\ (TX)} & \makecell[c]{\textbf{Tip-X}\\ (TX++)} & \textbf{APE} & \textbf{APE++} & \makecell[c]{$\Delta$ (TA++, TA)} & \makecell[c]{$\Delta$ (TX++, TX)} & \makecell[c]{$\Delta$ (TA++,TX)} & $\Delta$ (APE++,APE)\\
    \midrule 
    % \fontsize{1pt}{1pt}
EuroSAT & 1 & 48.383 & 63.288 & \textbf{68.259} & 63.597 & \textbf{68.527} & 65.901 &  \textbf{68.465} & 4.971 & 4.93 & 4.663  & 2.564 \\
EuroSAT & 2 & 48.383 & 68.267 & \textbf{72.292} & 68.576 & \textbf{73.012} & 71.14 &  \textbf{72.877} & 4.025 & 4.436 & 3.716  & 1.737 \\
EuroSAT & 4 & 48.383 & 73.354 & \textbf{74.683} & 73.547 & \textbf{75.041} & \textbf{75.802} &  75.292 & 1.329 & 1.494 & 1.136  & -0.51 \\
EuroSAT & 8 & 48.383 & 75.008 & \textbf{77.658} & 75.342 & \textbf{78.457} & \textbf{78.095} &  77.802 & 2.65 & 3.115 & 2.317  & -0.293 \\
EuroSAT & 16 & 48.383 & 78.852 & \textbf{81.407} & 78.864 & \textbf{81.782} & \textbf{81.494} &  81.387 & 2.556 & 2.918 & 2.543  & -0.107 \\
StanfordCars & 1 & 65.514 & 67.367 & \textbf{68.379} & 69.071 & \textbf{69.68} & 68.478 &  \textbf{69.22} & 1.011 & 0.609 & -0.692  & 0.742 \\
StanfordCars & 2 & 65.514 & 68.341 & \textbf{70.522} & 70.758 & \textbf{71.683} & 70.489 &  \textbf{71.77} & 2.18 & 0.924 & -0.236  & 1.281 \\
StanfordCars & 4 & 65.514 & 70.862 & \textbf{72.997} & 73.221 & \textbf{74.688} & 72.935 &  \textbf{74.07} & 2.135 & 1.467 & -0.224  & 1.135 \\
StanfordCars & 8 & 65.514 & 72.988 & \textbf{76.529} & 75.579 & \textbf{77.465} & 75.671 &  \textbf{77.063} & 3.54 & 1.886 & 0.949  & 1.392 \\
StanfordCars & 16 & 65.514 & 75.347 & \textbf{79.306} & 77.752 & \textbf{80.201} & 78.208 &  \textbf{80.496} & 3.959 & 2.45 & 1.555  & 2.288 \\
PLANTDOC & 1 & 34.994 & 39.78 & \textbf{39.888} & 40.384 & \textbf{41.138} & 41.117 &  \textbf{41.721} & 0.108 & 0.755 & -0.496  & 0.604 \\
PLANTDOC & 2 & 34.994 & 43.208 & \textbf{44.912} & 44.373 & \textbf{45.796} & 45.127 &  \textbf{47.348} & 1.703 & 1.423 & 0.539  & 2.221 \\
PLANTDOC & 4 & 34.994 & 46.766 & \textbf{49.051} & 47.003 & \textbf{49.59} & 49.116 &  \textbf{50.949} & 2.285 & 2.587 & 2.048  & 1.833 \\
PLANTDOC & 8 & 34.994 & 52.695 & \textbf{56.317} & 53.04 & \textbf{56.511} & 56.037 &  \textbf{58.905} & 3.622 & 3.471 & 3.277  & 2.868 \\
PLANTDOC & 16 & 34.994 & 56.425 & \textbf{61.082} & 56.231 & \textbf{61.427} & 61.751 &  \textbf{64.338} & 4.657 & 5.196 & 4.851  & 2.587 \\
DescribableTextures & 1 & 43.972 & 51.596 & \textbf{53.034} & 53.113 & \textbf{53.684} & 54.039 &  \textbf{54.59} & 1.438 & 0.571 & -0.079  & 0.551 \\
DescribableTextures & 2 & 43.972 & 54.886 & \textbf{56.994} & 56.462 & \textbf{57.289} & 58.747 &  \textbf{59.18} & 2.108 & 0.827 & 0.532  & 0.433 \\
DescribableTextures & 4 & 43.972 & 57.821 & \textbf{60.835} & 59.299 & \textbf{61.032} & \textbf{63.16} &  62.549 & 3.014 & 1.734 & 1.537  & -0.611 \\
DescribableTextures & 8 & 43.972 & 63.672 & \textbf{66.135} & 64.756 & \textbf{66.056} & \textbf{66.903} &  66.745 & 2.463 & 1.3 & 1.379  & -0.158 \\
DescribableTextures & 16 & 43.972 & 65.406 & \textbf{67.612} & 66.43 & \textbf{67.691} & \textbf{69.208} &  68.341 & 2.206 & 1.261 & 1.182  & -0.867 \\
StanfordDogs & 1 & 59.117 & 59.749 & \textbf{60.461} & 61.596 & \textbf{61.636} & 60.261 &  \textbf{61.028} & 0.712 & 0.04 & -1.136  & 0.767 \\
StanfordDogs & 2 & 59.117 & 60.317 & \textbf{61.368} & 62.796 & \textbf{62.92} & 61.408 &  \textbf{63.148} & 1.052 & 0.124 & -1.428  & 1.74 \\
StanfordDogs & 4 & 59.117 & 60.917 & \textbf{62.708} & 64.539 & \textbf{64.999} & 62.696 &  \textbf{65.659} & 1.791 & 0.46 & -1.831  & 2.963 \\
StanfordDogs & 8 & 59.117 & 62.54 & \textbf{64.971} & 67.302 & \textbf{67.734} & 65.327 &  \textbf{67.374} & 2.431 & 0.432 & -2.331  & 2.047 \\
StanfordDogs & 16 & 59.117 & 63.436 & \textbf{67.414} & 68.706 & \textbf{69.902} & 66.827 &  \textbf{69.742} & 3.979 & 1.196 & -1.292  & 2.915 \\
SUN397 & 1 & 62.579 & 65.529 & \textbf{66.713} & 66.584 & \textbf{67.058} & 66.687 &  \textbf{67.453} & 1.184 & 0.474 & 0.128  & 0.766 \\
SUN397 & 2 & 62.579 & 67.332 & \textbf{68.516} & 68.37 & \textbf{69.093} & 68.608 &  \textbf{69.602} & 1.184 & 0.723 & 0.146  & 0.994 \\
SUN397 & 4 & 62.579 & 68.791 & \textbf{70.35} & 70.025 & \textbf{70.929} & 70.94 &  \textbf{71.8} & 1.559 & 0.904 & 0.325  & 0.86 \\
SUN397 & 8 & 62.579 & 70.441 & \textbf{71.781} & 71.753 & \textbf{72.809} & 72.571 &  \textbf{72.895} & 1.34 & 1.055 & 0.028  & 0.324 \\
SUN397 & 16 & 62.579 & 71.635 & \textbf{72.874} & 72.955 & \textbf{73.776} & \textbf{73.429} &  73.332 & 1.239 & 0.821 & -0.081  & -0.097 \\
FGVCAircraft & 1 & 24.752 & 28.363 & \textbf{29.033} & 29.573 & \textbf{30.253} & 28.833 &  \textbf{29.163} & 0.67 & 0.68 & -0.54  & 0.33 \\
FGVCAircraft & 2 & 24.752 & 29.173 & \textbf{29.983} & 31.383 & \textbf{31.523} & 30.223 &  \textbf{31.013} & 0.81 & 0.14 & -1.4  & 0.79 \\
FGVCAircraft & 4 & 24.752 & 32.593 & \textbf{34.063} & 34.653 & \textbf{35.914} & 33.773 &  \textbf{35.274} & 1.47 & 1.26 & -0.59  & 1.501 \\
FGVCAircraft & 8 & 24.752 & 35.934 & \textbf{37.424} & 37.954 & \textbf{38.344} & 38.384 &  \textbf{39.434} & 1.49 & 0.39 & -0.53  & 1.05 \\
FGVCAircraft & 16 & 24.752 & 39.774 & \textbf{41.504} & 41.164 & \textbf{42.424} & 42.084 &  \textbf{42.384} & 1.73 & 1.26 & 0.34  & 0.3 \\
OxfordPets & 1 & 89.071 & 89.697 & \textbf{90.588} & 90.424 & \textbf{90.851} & 91.242 &  \textbf{91.387} & 0.89 & 0.427 & 0.164  & 0.145 \\
OxfordPets & 2 & 89.071 & 90.006 & \textbf{90.96} & 91.133 & \textbf{91.705} & 91.76 &  \textbf{92.205} & 0.954 & 0.572 & -0.173  & 0.445 \\
OxfordPets & 4 & 89.071 & 90.388 & \textbf{91.633} & 91.496 & \textbf{92.087} & 91.642 &  \textbf{92.105} & 1.245 & 0.591 & 0.136  & 0.463 \\
OxfordPets & 8 & 89.071 & 90.77 & \textbf{92.241} & 92.141 & \textbf{92.686} & 91.923 &  \textbf{92.332} & 1.472 & 0.545 & 0.1  & 0.409 \\
OxfordPets & 16 & 89.071 & 91.051 & \textbf{92.414} & 92.65 & \textbf{93.05} & 92.214 &  \textbf{92.269} & 1.363 & 0.4 & -0.236  & 0.055 \\
CUB & 1 & 55.009 & 59.318 & \textbf{60.301} & 61.103 & \textbf{61.995} & 59.437 &  \textbf{59.993} & 0.983 & 0.892 & -0.802  & 0.556 \\
CUB & 2 & 55.009 & 61.514 & \textbf{62.128} & 63.536 & \textbf{64.457} & 62.92 &  \textbf{63.312} & 0.614 & 0.92 & -1.408  & 0.392 \\
CUB & 4 & 55.009 & 64.652 & \textbf{65.781} & 67.127 & \textbf{68.57} & 66.681 &  \textbf{67.172} & 1.129 & 1.443 & -1.346  & 0.491 \\
CUB & 8 & 55.009 & 68.41 & \textbf{69.177} & 70.961 & \textbf{71.415} & 70.178 &  \textbf{70.342} & 0.767 & 0.453 & -1.785  & 0.164 \\
CUB & 16 & 55.009 & 71.798 & \textbf{72.823} & 72.711 & \textbf{74.238} & 74.33 &  \textbf{74.345} & 1.025 & 1.527 & 0.112  & 0.015 \\
ImageNet & 1 & 68.804 & 69.28 & \textbf{69.536} & 69.389 & \textbf{69.568} & 69.493 &  \textbf{69.822} & 0.256 & 0.179 & 0.147  & 0.329 \\
ImageNet & 2 & 68.804 & 69.477 & \textbf{69.805} & 69.509 & \textbf{69.812} & 69.804 &  \textbf{70.289} & 0.328 & 0.303 & 0.297  & 0.485 \\
ImageNet & 4 & 68.804 & 69.791 & \textbf{70.359} & 69.864 & \textbf{70.359} & 70.247 &  \textbf{70.845} & 0.569 & 0.495 & 0.495  & 0.598 \\
ImageNet & 8 & 68.804 & 70.249 & \textbf{70.949} & 70.459 & \textbf{71.012} & 70.81 &  \textbf{71.367} & 0.699 & 0.553 & 0.489  & 0.557 \\
ImageNet & 16 & 68.804 & 70.753 & \textbf{71.505} & 70.973 & \textbf{71.587} & 71.094 &  \textbf{71.811} & 0.753 & 0.613 & 0.532  & 0.717 \\
Caltech101 & 1 & 93.306 & 93.563 & \textbf{93.874} & 93.414 & \textbf{93.739} & 93.671 &  \textbf{94.32} & 0.311 & 0.325 & 0.46  & 0.649 \\
Caltech101 & 2 & 93.306 & 93.969 & \textbf{94.469} & 94.145 & \textbf{94.442} & 94.51 &  \textbf{94.794} & 0.5 & 0.297 & 0.325  & 0.284 \\
Caltech101 & 4 & 93.306 & 94.388 & \textbf{94.929} & 93.942 & \textbf{94.97} & 94.861 &  \textbf{95.024} & 0.541 & 1.028 & 0.987  & 0.163 \\
Caltech101 & 8 & 93.306 & 94.686 & \textbf{95.159} & 94.983 & \textbf{95.186} & 94.943 &  \textbf{95.402} & 0.473 & 0.203 & 0.176  & 0.459 \\
Caltech101 & 16 & 93.306 & 94.97 & \textbf{95.456} & 95.01 & \textbf{95.659} & 95.078 &  \textbf{95.483} & 0.487 & 0.649 & 0.446  & 0.405 \\
Food101 & 1 & 85.888 & \textbf{85.986} & 85.96 & 85.955 & \textbf{85.998} & \textbf{86.044} &  86.025 & -0.025 & 0.043 & 0.006  & -0.019 \\
Food101 & 2 & 85.888 & \textbf{86.133} & 86.086 & 86.178 & \textbf{86.238} & 86.196 &  \textbf{86.2} & -0.047 & 0.059 & -0.092  & 0.004 \\
Food101 & 4 & 85.888 & \textbf{86.232} & 86.134 & \textbf{86.238} & 86.21 & \textbf{86.403} &  86.261 & -0.098 & -0.028 & -0.103  & -0.142 \\
Food101 & 8 & 85.888 & 86.194 & \textbf{86.251} & 86.375 & \textbf{86.387} & \textbf{86.461} &  86.369 & 0.057 & 0.012 & -0.124  & -0.092 \\
Food101 & 16 & 85.888 & \textbf{86.43} & 86.394 & 86.517 & \textbf{86.565} & \textbf{86.743} &  86.432 & -0.036 & 0.048 & -0.123  & -0.311 \\
UCF101 & 1 & 67.46 & 71.716 & \textbf{72.024} & 72.553 & \textbf{72.667} & \textbf{73.187} &  73.055 & 0.308 & 0.115 & -0.529  & -0.132 \\
UCF101 & 2 & 67.46 & 73.777 & \textbf{73.857} & 75.17 & \textbf{75.24} & \textbf{76.835} &  75.443 & 0.079 & 0.07 & -1.313  & -1.392 \\
UCF101 & 4 & 67.46 & \textbf{74.007} & 73.795 & \textbf{75.399} & 75.17 & \textbf{76.853} &  75.178 & -0.211 & -0.229 & -1.604  & -1.675 \\
UCF101 & 8 & 67.46 & \textbf{77.284} & 76.509 & 78.298 & \textbf{78.377} & \textbf{79.135} &  77.487 & -0.775 & 0.079 & -1.789  & -1.648 \\
UCF101 & 16 & 67.46 & \textbf{78.421} & 77.602 & 78.773 & \textbf{79.038} & \textbf{79.637} &  78.395 & -0.819 & 0.264 & -1.172  & -1.242 \\
OxfordFlowers & 1 & 70.767 & \textbf{83.435} & 82.961 & \textbf{84.504} & 84.193 & \textbf{87.468} &  85.099 & -0.474 & -0.311 & -1.543  & -2.369 \\
OxfordFlowers & 2 & 70.767 & \textbf{87.319} & 86.615 & \textbf{88.415} & 87.86 & \textbf{91.122} &  88.578 & -0.704 & -0.555 & -1.8  & -2.544 \\
OxfordFlowers & 4 & 70.767 & \textbf{90.378} & 89.078 & \textbf{91.135} & 90.472 & \textbf{93.247} &  90.797 & -1.299 & -0.663 & -2.057  & -2.45 \\
OxfordFlowers & 8 & 70.767 & \textbf{92.719} & 91.487 & \textbf{92.922} & 92.57 & \textbf{94.248} &  93.166 & -1.232 & -0.352 & -1.435  & -1.082 \\
OxfordFlowers & 16 & 70.767 & \textbf{94.262} & 92.732 & \textbf{94.546} & 93.341 & \textbf{95.886} &  95.168 & -1.529 & -1.204 & -1.814  & -0.718 \\
\midrule
Average fine-grained & 1 & 60.979 & 65.649 & \textbf{66.613} & 66.612 & \textbf{67.427} & 66.954 &  \textbf{67.365} & 0.963 & 0.815 & 0.0  & 0.411 \\
Average fine-grained & 2 & 60.979 & 67.558 & \textbf{68.757} & 68.887 & \textbf{69.72} & 69.422 &  \textbf{70.039} & 1.2 & 0.834 & -0.13  & 0.617 \\
Average fine-grained & 4 & 60.979 & 69.85 & \textbf{71.081} & 71.109 & \textbf{72.143} & 71.847 &  \textbf{72.287} & 1.231 & 1.034 & -0.028  & 0.44 \\
Average fine-grained & 8 & 60.979 & 72.329 & \textbf{73.941} & 73.76 & \textbf{74.801} & 74.379 &  \textbf{74.994} & 1.612 & 1.041 & 0.181  & 0.615 \\
Average fine-grained & 16 & 60.979 & 74.341 & \textbf{76.195} & 75.507 & \textbf{76.935} & 76.711 &  \textbf{77.308} & 1.854 & 1.427 & 0.688  & 0.597 \\
Average all & 1 & 62.115 & 66.333 & \textbf{67.215} & 67.233 & \textbf{67.928} & 67.561 &  \textbf{67.953} & 0.882 & 0.695 & -0.018  & 0.392 \\
Average all & 2 & 62.115 & 68.123 & \textbf{69.179} & 69.343 & \textbf{70.076} & 69.921 &  \textbf{70.411} & 1.056 & 0.733 & -0.164  & 0.49 \\
Average all & 4 & 62.115 & 70.067 & \textbf{71.171} & 71.249 & \textbf{72.145} & 72.025 &  \textbf{72.355} & 1.104 & 0.896 & -0.078  & 0.33 \\
Average all & 8 & 62.115 & 72.399 & \textbf{73.756} & 73.705 & \textbf{74.644} & 74.335 &  \textbf{74.763} & 1.357 & 0.939 & 0.052  & 0.428 \\
Average all & 16 & 62.115 & 74.183 & \textbf{75.723} & 75.235 & \textbf{76.477} & 76.284 &  \textbf{76.709} & 1.541 & 1.243 & 0.489  & 0.425 \\
    \bottomrule
    \end{tabular}}
    \vspace{2mm}
    \caption{Average results by number of shots over 3 seeds including APE.}
    \label{table:table8}
\end{table}

\end{document}